\documentclass{article}

\usepackage{microtype}
\usepackage{graphicx}
\usepackage{booktabs} 

\usepackage{hyperref}

\usepackage[accepted]{icml2025}

\usepackage{amsmath}
\usepackage{amssymb}
\usepackage{mathtools}
\usepackage{amsthm}

\usepackage{algorithmic}
\usepackage{xcolor}
\usepackage{colortbl}

\usepackage{float}
\usepackage{wrapfig}
\usepackage{caption}
\usepackage{subcaption}
\usepackage{graphicx}
\usepackage{multirow}
\usepackage{bbm}
\usepackage{booktabs}

\usepackage[capitalize,noabbrev]{cleveref}

\theoremstyle{plain}
\newtheorem{theorem}{Theorem}[section]
\newtheorem{proposition}[theorem]{Proposition}

\theoremstyle{definition}

\newtheorem{assumption}[theorem]{Assumption}
\theoremstyle{remark}
\newtheorem{remark}[theorem]{Remark}
\definecolor{Gray}{gray}{0.95}

\newcommand{\atom}{\mathbf{D}}
\newcommand{\coeff}{\boldsymbol{\alpha}}

\newcommand{\localw}{\mathbf{w}_k}
\newcommand{\globalw}{\mathbf{w}}
\newcommand{\fullw}{\mathbf{\hat{w}}}
\newcommand{\param}{\boldsymbol{\theta}}
\newcommand{\extractor}{\boldsymbol{\phi}}

\usepackage[textsize=tiny]{todonotes}

\icmltitlerunning{Extra Clients at No Extra Cost}

\begin{document}

\twocolumn[
\icmltitle{Extra Clients at No Extra Cost: Overcome Data Heterogeneity in Federated Learning with Filter Decomposition}




\begin{icmlauthorlist}
\icmlauthor{Wei Chen}{pd}
\icmlauthor{Qiang Qiu}{pd}
\end{icmlauthorlist}

\icmlaffiliation{pd}{Purdue University, West Lafayette, IN, USA}

\icmlcorrespondingauthor{Wei Chen}{chen2732@purdue.edu}

\icmlkeywords{Machine Learning, ICML}

\vskip 0.3in
]



\printAffiliationsAndNotice{}  

\begin{abstract}
Data heterogeneity is one of the major challenges in federated learning (FL), which results in substantial client variance and slow convergence. 
In this study, we propose a novel solution: 
decomposing a convolutional filter in FL into a linear combination of filter subspace elements, \textit{i.e.}, filter atoms. 
This simple technique transforms global filter aggregation in FL into 
aggregating filter atoms and their atom coefficients.
The key advantage here involves mathematically \textit{generating numerous cross-terms} by expanding the product of two weighted sums from filter atom and atom coefficient. These cross-terms effectively emulate many additional \textit{latent clients}, significantly reducing model variance, which is validated by our theoretical analysis and empirical observation. 
Furthermore, our method permits different training schemes for filter atoms and atom coefficients for highly adaptive model personalization and communication efficiency. 
Empirical results on benchmark datasets demonstrate that our filter decomposition technique substantially improves the accuracy of FL methods, confirming its efficacy in addressing data heterogeneity.
\end{abstract}

\section{Introduction}
\label{intro}
Federated learning (FL) is a collaborative learning technique that aggregates models from local clients while ensuring data privacy~\citep{mcmahan2017communication}. This approach has demonstrated remarkable success in various application domains, such as wearable devices~\citep{nguyen2019diot}, medical diagnosis~\citep{dong2020can, yang2021flop}, and mobile phones~\citep{li2020federated}. 

The heterogeneity of data distribution across different clients poses a significant challenge to FL~\citep{mcmahan2017communication}. In many real-world applications, the data can be non-independent and identically distributed (non-IID) among clients, which adversely impacts the performance of federated learning. Variations in user behavior can result in heterogeneous data distributions. For instance, in face recognition tasks using user photos~\citep{adjabi2020past}, it is common to encounter significant disparities in facial appearances among different individuals captured in their respective pictures.
The variations in local data give rise to divergent local optima in contrast to the global optimum. 
This phenomenon results in a notable variance during the aggregation of local models and introduces potential challenges in attaining global convergence via local training procedures~\citep{li2020federated, mcmahan2017communication}.

\begin{figure}[t]
  \centering

 \begin{subfigure}{0.43\textwidth}
     \centering
     \includegraphics[trim={260pt 200pt 300pt 125pt}, clip, width=\textwidth]{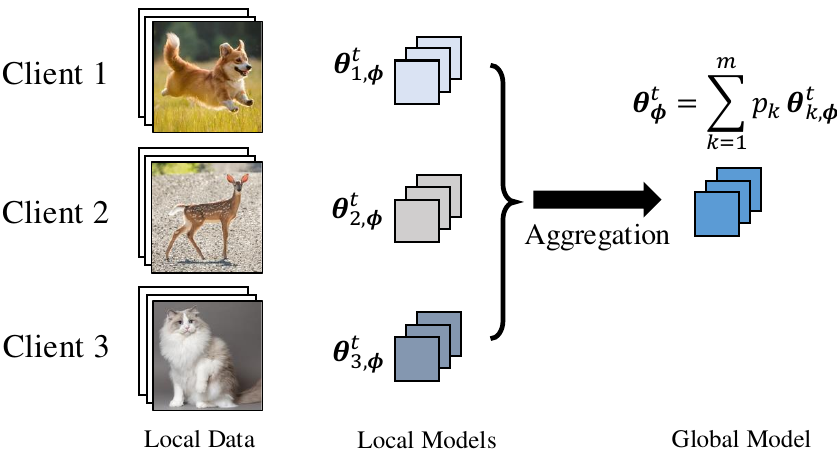} 
     \caption{ }
 \end{subfigure}

  \begin{subfigure}{0.43\textwidth}
    \centering
    \includegraphics[trim={220pt 130pt 250pt 35pt}, clip, width=\textwidth]{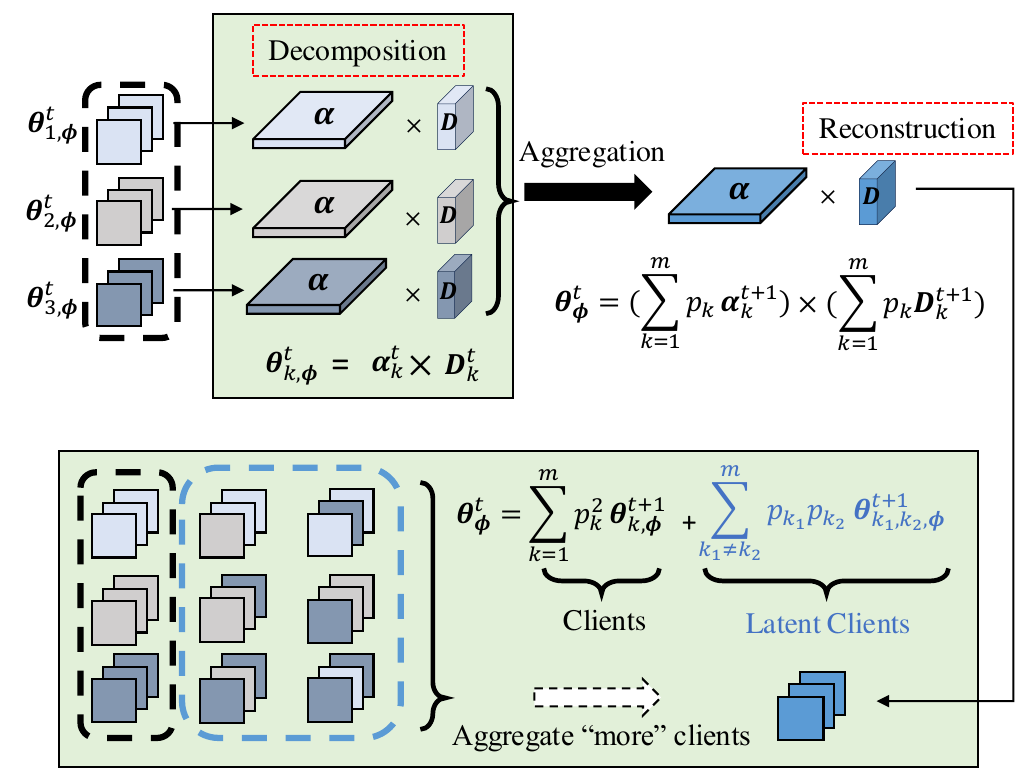} 
    \caption{}
  \end{subfigure}

  \caption{
  (a) The aggregation of convolutional filters, e.g., FedAvg~\cite{mcmahan2017communication}.
  (b) We decompose the convolutional filters as \textit{filter atoms} $\atom$ and \textit{atom coefficients} $\coeff$. 
  During the aggregation phase, we \textbf{separately} average the filter atoms $\atom$ and atom coefficients $\coeff$, and subsequently reconstruct the global model by multiplying the aggregated $\coeff$ and aggregated $\atom$. In contrast to conventional FL aggregation methods like FedAvg, this mathematical operation naturally leads to additional local model variants
  significantly reducing the variance of local updates, without introducing extra computation cost or communication overhead.
  }
  \label{fig:fl_sys_diagram}
\end{figure}

We observe that including more clients in the model aggregation process can lower training variance and address challenges related to non-IID data. Thus, we propose a method to create additional clients at no extra cost, starting with model decomposition, followed by global aggregation and model reconstruction. We achieve this by decomposing the filters in a convolutional neural network (CNN) into a linear combination of subspace components, \textit{i.e.}, filter atoms. Our system diagram in Figure~\ref{fig:fl_sys_diagram} depicts how this method facilitates the creation of global aggregated filters by multiplying the weighted sums of local filter atoms and their corresponding coefficients. It is easy to notice that expanding these weighted sums naturally leads to many cross-terms, which implicitly emulate numerous additional latent clients at no cost. As shown in our theoretical analysis and empirical obviation, these latent clients play a critical role in significantly lowering the variance of the overall model and speeding up the convergence during the training process.

The above filter decomposition further permits different training schemes for filter atoms and atom coefficients, which enables various opportunities to be explored in FL. For example, for \textit{model personalization}, we can have filter atoms potentially focus more on personalized local knowledge, while the decomposed atom coefficients capture more shared knowledge about the combination rules of filter atoms. 
This strategy is in harmony with task subspace modeling principles, which suggest that task parameters occupy a low-dimensional subspace, allowing tasks to be represented as combinations of latent basis tasks. This concept has been explored in various studies~\citep{chen2024inner, chenggraph, evgeniou2007multi, kumar2012learning, lezama2021run, maurer2013sparse, miao2021spatiotemporal, miao2021continual, romera2013multilinear, wang2021image, zhang2021survey}.
With this concept, our method becomes highly adaptable to personalized federated learning. 
For \emph{communication reduction}, we can adopt different update frequencies for filter atoms and atom coefficients. The communication overhead can be minimized by prioritizing the transmission of filter atoms over atom coefficients since filter atoms usually only contain a few hundred parameters. 

We evaluate the effectiveness of our approach on standard classification benchmark datasets. Our results demonstrate that it can be seamlessly integrated with various federated learning methods, resulting in enhanced accuracy. Additionally, we provide convergence analyses of our proposed approach and investigate the training trajectory in the loss landscape to visualize the advantages of our method.

We summarize our contributions as follows,
\begin{itemize}
    \item We propose a highly simple yet effective scheme to handle data heterogeneity in FL by decomposing convolutional filters into filter atoms and atom coefficients.
    \item The mathematical formulation of the proposed approach naturally results in numerous additional local model variants as latent clients.
    \item We show both theoretically and empirically that those implicitly introduced local model variants result in a reduction in the variance of the global model and lead to faster convergence.
    \item Our approach enables various opportunities to be explored in FL, such as model personalization and communication reduction.
\end{itemize}

\section{Related Works}
\label{relates}
\paragraph{Data Heterogeneity.}
Various strategies have been proposed to enhance the global model accuracy of federated learning (FL) in the presence of heterogeneous data. One approach is data-based, which aims to address the issue of client drift by reducing the statistical heterogeneity among the data stored on clients~\citep{yoonfedmix, zhao2018federated}. Alternatively, model-based methods aim to preserve valuable information related to the inherent diversity of client behaviors. Such methods aim to learn a robust global FL model that can be personalized for each individual client in the future or to improve the adaptation performance of the local model~\citep{acarfederated, karimireddy2020scaffold, li2021model, li2020federated2, mu2023fedproc}. Reducing aggregation variance can effectively lead to improved convergence, resulting in an enhanced global model~\citep{chen2021fedmax, jhunjhunwala2022fedvarp, li2023effectiveness, malinovsky2022variance}.
In contrast to the strategy of personalizing a well-trained global model, recent research has explored the use of personalized federated learning (pFL) approaches, in which personalized models are trained for each client~\citep{achituve2021personalized, bhardwaj2020new, duan2021flexible, ghosh2020efficient, huang2021personalized, sattler2020clustered, sharma2021tesseract, sharma2023learn, sharma2023flair, zhangpersonalized}. Various methods have been proposed for pFL, such as meta-learning~\citep{acar2021debiasing, fallah2020personalized, khodak2019adaptive}, multi-task learning~\citep{li2021ditto, smith2017federated, t2020personalized}, and parameter decoupling~\citep{fedper, collins2021exploiting, hyeonfedpara, fedpac, lgfedavg}.

\paragraph{Parameter Decoupling.}
Among pFL methods, parameter decoupling seeks to achieve personalization by separating the local parameters from the global FL model parameters, such that private parameters are only trained on local client data and are not shared with the FL server. This allows for the learning of task-specific representations, resulting in improved personalization. FedPer~\citep{fedper}, FedRep~\citep{collins2021exploiting}, and FedPAC~\citep{fedpac} introduce algorithms for training local heads and a global network body, with their primary distinction lying in their respective approaches to leveraging global knowledge. LG-FedAvg~\citep{lgfedavg} proposed a representation learning method that attempts to learn many local representations and a single global head. FedPara\citep{hyeonfedpara} re-parameterizes weight parameters of layers using low-rank weights followed by the Hadamard product and achieves personalization by separating the roles of each sub-matrix into global and local inner matrices. 
In contrast, our method utilizes filter decomposition, allowing the parameter decoupling inside each convolutional layer.

\section{Preliminary}
\label{formulation}

\paragraph{Federated learning formulation.}
Federated learning aims to solve the learning task without explicitly sharing local data. During the training time, a central server coordinates the global learning across a network, where each node is a device with local data and performs a local learning task. The client $k$ contains its own data distribution $P^{(k)}_{XY}$ on $\mathcal{X} \times \mathcal{Y}$, where $\mathcal{X}$ is the input space and $\mathcal{Y}$ is the label space. The objective of FL~\citep{mcmahan2017communication} is to minimize:
    \begin{equation}
     \underset{\globalw}{\mathrm{min}} \ \ \  F\left(\globalw \right)=\sum_{k=1}^{M} p_k \cdot F_k(\localw),
    \label{eq:global_object}
    \end{equation}
where $\globalw$ is the parameters of the global model, $F_k(\localw)$ is the local objective at client $k$ which is typically the loss function with model parameters $\localw$; 
$M$ is the total number of devices. 
Given $m = C \cdot M$ is the number of devices selected at any given communication round, where $C$ is the proportion of selected devices, $n_k$ as the number of samples available at the device $k$ and $n=\sum_{k=1}^{m} n_k$ as the total number of samples on selected devices, we have $\sum_{k=1}^{m} p_k = 1$, $p_k=\frac{n_k}{n}$. The local objective $F_k(\localw)$ at client is further defined by:

\begin{equation}
     F_k(\localw) = \frac{1}{n_k} \sum_{j=1}^{n_k} \mathcal{L}_{(x,y) \sim P^{(k)}_{XY}}(\localw; \mathbf{x}_j, y_j),
    \label{eq:local_object}
\end{equation}
where $\mathcal{L}(\cdot;\cdot)$ is a client-specific loss function, e.g., cross-entropy loss; and $\mathbf{x}_j \in \mathcal{X}$ is the input data, $y_j \in \mathcal{Y}$ is the corresponding label.

\paragraph{More involved clients improve convergence.}
Previous works~\cite{karimireddy2020scaffold, liconvergence} show that increasing the number of participating clients per communication round generally improves convergence by reducing gradient estimation variance and enhancing stability, as a larger aggregation leads to a better approximation of the global gradient. The convergence rate typically scales linearly as $\mathcal{O}(m)$. However, in non-IID settings, data heterogeneity limits the effectiveness of involving more clients due to client drift~\cite{karimireddy2020scaffold, liconvergence}. Additionally, a higher number of clients increases communication overhead, reducing overall communication efficiency.

\section{Proposed Approach}
\label{sec:approach}

Prior research has explored various techniques to decrease variance during model aggregation. In our approach, we achieve reduced variance by introducing additional clients, referred to as \textit{latent clients}. 
We formally define this process and demonstrate that such a formulation effectively reduces variance and aids in handling non-IID data. Additionally, incorporating latent clients allows for an increase in the number of participating clients without adding extra communication overhead.

We first decouple the model $\globalw$ as feature extractor $\extractor: \mathbb{R}^{h\times w \times c} \rightarrow \mathbb{R}^d$, which is a learnable network parameterized by $\param_{\extractor}$ and maps data to a $d$-dimensional feature space, and heads $\mathbf{h}: \mathbb{R}^d \rightarrow \mathcal{Y}$, which are parameterized by $\param_{h}$ and maps features to the label space. $h \times w$ is the shape of input image, $c$ is the number of input channels. 

We then formulate our filter decomposition method based on~\citep{dcf}, which involves decomposing each convolutional layer of the feature extractor $\extractor$ into two standard convolutional layers: a \textit{filter atom} layer that models filter subspace, and an \textit{atom coefficient} layer with $1 \times 1$ filters that model combination rules of filter atoms. Specifically, the convolutional filter $\mathcal{F} \in \mathbb{R}^{c'\times c\times k_a\times k_a}$ is decomposed over $m_a$ filter subspace elements, \textit{i.e.}, filter atoms $\atom \in \mathbb{R}^{m_a \times k_a\times k_a}$, linearly combined by atom coefficients $\coeff \in \mathbb{R}^{c'\times c \times m_a}$, where $c'$ and $c$ are the numbers of input and output channels, $k_a$ is the kernel size. Convolutional filters are the dominant subset of parameters of the feature extractor, $\mathcal{F} \subseteq \param_{\extractor}$ and the feature extractor $\extractor$ comprises multiple convolutional layers in practice, but we simplify the notations by setting $\param_{\extractor}=\mathcal{F}=\coeff \times \atom$
. 

The filter decomposition is illustrated in Figure~\ref{fig:fl_sys_diagram}(b). Each convolutional filter $\param_{\extractor, ij} \in \mathbb{R}^{k_a\times k_a}$ is constructed by the combination of filter atoms $\param_{\extractor, ij} = \sum_{q=1}^{m_a} \alpha_{ijq} \atom_{q}$, where $\atom_{q} \in \mathbb{R}^{k_a\times k_a}$ is the one filter atom and $\alpha_{ijq}$ is the element in $\coeff$.
With the above formulation, the local objective becomes,
\begin{align}
     F_k(\localw) & = F_k(\coeff_k, \atom_k, \param_{k, h}) \nonumber \\
     & = \frac{1}{n_k} \sum_{j=1}^{n_k} \mathcal{L}(\coeff_k, \atom_k, \param_{k, h}; \mathbf{x}_j, y_j),
    \label{eq:global_object}
\end{align}
where $\localw = \{\coeff_k, \atom_k, \param_{k, h}\}$.
The majority of the training process remains consistent with FL while the sole distinction lies in the aggregation and reconstruction step.

\paragraph{Local Training.} 
We perform parameter updates using the gradients of the loss function. Here, we write the update of each part explicitly, 
\begin{equation}
    \begin{bmatrix} \coeff^{t+1}_k \\ \atom^{t+1}_k \\ \param_{k, h}^{t+1} \end{bmatrix} \xleftarrow{} 
    \begin{bmatrix}\coeff^{t}_k - \eta_t \nabla_{\coeff^{t}_k} F_k \\ \atom^{t}_k - \eta_t \nabla_{\atom^{t}_k} F_k \\ \param_{k, h}^{t} - \eta_t \nabla_{\param_{k, h}^{t}} F_k \end{bmatrix}, 
\end{equation}

\paragraph{Global Aggregation.}
The model separately aggregates the $\coeff$, $\atom$, and $\param_h$,
\begin{equation}
    \begin{bmatrix} 
        \coeff^{t+1} \\ 
        \atom^{t+1} \\ 
        \param_{h}^{t+1} 
    \end{bmatrix} 
    \xleftarrow{} 
    \begin{bmatrix} 
        \sum_{k=1}^{m} \frac{n_k}{n} \coeff^{t+1}_k \\ 
        \sum_{k=1}^{m} \frac{n_k}{n} \atom^{t+1}_k \\ 
        \sum_{k=1}^{m} \frac{n_k}{n}  \param_{k, h}^{t+1} 
    \end{bmatrix}.
    \label{eq:dcf_weight_coeff_update}
\end{equation}


\paragraph{Global Reconstruction.} The global convolutional filter is then formed by multiplying $\coeff^{t+1}$ and $\atom^{t+1}$ of selected $m$ clients, \textit{i.e.}, 
\begin{equation}
    \param_{\extractor}^{t+1} \xleftarrow{} \coeff^{t+1} \times \atom^{t+1}.
    \label{eq:dcf_weight_overall_update}
\end{equation} 

And a new global model has parameters $\globalw^{t+1}=\{\param_{\extractor}^{t+1}, \param_{h}^{t+1}\}$ for the next round local update. In practice, the reconstruction step is achieved by the neural network design without incurring any additional computational overhead. The algorithm is summarized in Appendix Algorithm~\ref{alg:algorithm}.



\subsection{Personalization}


    

This formulation enables model personalization for FL. Each client maintains both local filter atoms $\atom_{l}$ and global filter atoms $\atom_{g}$. The training procedure for global filter atoms adheres to the federated learning update rule explained in the preceding sections. In contrast, the local filter atoms undergo only local training without any global aggregation. Specifically, the local filter atoms update locally using the local data with the \emph{fixed} atom coefficients at round $t$, expressed as,
\begin{align*}
    \atom^{t+1}_{l,k} = \atom^{t}_{l,k} - \eta_t \nabla_{\atom^{t}_{l,k}} F_k,
\end{align*}
where $\atom^{t}_{l,k}$ is the local filter atoms of client $k$ at communication round $t$, and $\nabla_{\atom^{t}_{l,k}} F_k$ denotes the gradient of the local loss function concerning the local filter atoms.


In this formulation, the global combination rule is communicated via atom coefficients $\coeff^t$. New atom coefficients $\coeff^{t+1}$ are acquired through aggregation from selected local clients, $\coeff^{t+1}= \sum_{k=1}^{m} \frac{n_k}{n} \coeff^{t+1}_k$. Atom coefficients can be interpreted as shared knowledge for combining filter atoms. Throughout the local training process, the fixed atom coefficients function as guides, directing the local filter atoms $\atom^{t+1}_{l,k}$ to adapt to specific representations of the local data while maintaining an awareness of the global coefficients. 

\subsection{Imbalance Updates of $\coeff$ and $\atom$}
This formulation also provides a strategy to reduce the communication overhead in FL. Compared with atom coefficients $\coeff \in \mathbb{R}^{m_a\times c'\times c}$, the filter atoms $\atom \in \mathbb{R}^{k_a\times k_a \times m_a}$ have significantly fewer parameters since $k_a \times k_a \ll c'\times c$. To incorporate this finding, we further adopt a fast/slow communication protocol, which prioritizes the transmission of $\atom$ over $\coeff$ to minimize communication costs. More precisely, we introduce a parameter $\beta$, that determines the frequency of atom coefficient communication. For instance, if $\beta=1/5$, the atom coefficients are communicated and updated once every five rounds.
With this, the atom coefficients of the model can be aggregated as follows, 
\begin{equation}
    \coeff^{t+1} \xleftarrow{} \sum_{k=1}^{m} \frac{n_k}{n} \coeff^{t+1}_k \mathbf{\delta}_{\{\beta t \in \mathbb{N}\}} + \coeff^{t} \mathbf{\delta}_{\{\beta t \notin \mathbb{N}\}},
    \label{eq:beta_dcf_weight_coeff_update}
\end{equation}
where $\mathbf{\delta}_{\{\beta t \in \mathbb{N}\}}$ is the indicator which equals to 1 only when $\beta t$ is an integer. The algorithm is summarized in Appendix Algorithm~\ref{alg:algorithm_fast_slow}.

\section{Analysis}
In this section, we provide theoretical analyses of the proposed approach with regard to convergence, and the impact of filter decomposition on reducing global model variance.

\subsection{Property of Latent Clients}
Based on our formulation, the reconstruction of decomposed filters results in a natural incorporation of additional local model variants. By inserting (\ref{eq:dcf_weight_coeff_update}) into (\ref{eq:dcf_weight_overall_update}), we have
\begin{align}
    \param_{\extractor}^{t+1} & = (\sum_{k=1}^{m} \frac{n_k}{n} \coeff^{t+1}_k) \times (\sum_{k=1}^{m} \frac{n_k}{n} \atom^{t+1}_k) \nonumber \\
            & = \underbrace{\sum_{k=1}^{m} \frac{n_k^2}{n^2} \param_{k, \extractor}^{t+1}}_{\text{Clients}} + \underbrace{\color{blue}{\sum_{k_1 \neq k_2}^{m} \frac{n_{k_1} \cdot n_{k_2}}{n^2} \param_{k_1, k_2, \extractor}^{t+1}}}_{\text{Latent Clients}},
    \label{eq:dcf_weight_update}
\end{align} 
where $\param_{k, \extractor}^{t+1}=\coeff^{t+1}_k \times \atom^{t+1}_k$ and $\param_{k_1, k_2, \extractor}^{t+1}=\coeff^{t+1}_{k_1} \times \atom^{t+1}_{k_2}$.
Compared with the weight update of FedAvg, (\ref{eq:dcf_weight_update}) contains both averaging of selected clients represented in the first term and extra reconstructed latent clients in the second term, as illustrated in Figure~\ref{fig:fl_sys_diagram}(b).

The latent clients $\param_{k_1, k_2, \extractor}^{t+1}$ are created through atom swapping, a formulation that provides stable representations for knowledge retention and new learning~\cite{miao2021continual}. Specifically, the loss of latent client $\param_{k_1, k_2, \extractor}$ is bounded, given the loss of client $\param_{k_1, \extractor}$ and $\param_{k_2, \extractor}$.

\subsection{Reduced Global Model Variance}
\label{sec:reduce_var}

The additional reconstructed local model variants contribute to a decrease in the variance of the aggregated global model, as shown next.

\begin{proposition}
    Consider $\param_{k, \extractor}$ and $\param_{k_1, k_2, \extractor}$ as independent random variables,
    the parameter obtained by methods without filter decomposition is $\param_{\extractor}=\sum_{k=1}^{m} \frac{n_k}{n} \param_{k, \extractor}$, and parameter obtained by (\ref{eq:dcf_weight_update}) is $\param_{\extractor}'=\sum_{k=1}^{m} \frac{n_k^2}{n^2} \param_{k, \extractor} + \sum_{k_1 \neq k_2}^{m} \frac{n_{k_1} \cdot n_{k_2}}{n^2} \param_{k_1, k_2, \extractor}$. We have
    \begin{align*}
        \mathbb{E} ||\param_{\extractor}' - \mathbb{E} (\param_{\extractor}')||^2 
        \leq & \mathbb{E} ||\param_{\extractor} - \mathbb{E} (\param_{\extractor})||^2.
    \end{align*}
\end{proposition}

We provide further details on the analysis in Appendix~\ref{theory:variance}. Empirical results are presented in Section~\ref{sec:more_clients} to demonstrate that increasing the number of clients involved in the aggregation process leads to a reduction in the variance of the aggregated model.

\subsection{Convergence Analysis}
The objective function of client $k$ is denoted by $F_{k}$, where $k=0,1,2,...m-1$. We assume the following properties of the objective function which are adapted from~\citep{liconvergence}:

\begin{assumption}
    \label{assp:lsmooth}
    $F_{k}$ are all L-smooth, that is, for all $v$ and $w$, $F_{k}(v)\leq F_{k}(w) + (v-w)^{T}\triangledown F_{k}(w) + \frac{L}{2}\Vert v-w \Vert^{2}_{2}$.
\end{assumption}

\begin{assumption}
    \label{assp:muconvex}
    $F_{k}$ are all $\mu$-strongly convex, that is, for all $v$ and $w$, $F_{k}(v)\geq F_{k}(w) + (v-w)^{T}\triangledown F_{k}(w) + \frac{\mu}{2}\Vert v-w \Vert^{2}_{2}$.
\end{assumption}



\begin{assumption}
    \label{assp:gradbound}
    The squared norm of gradients is uniformly bounded, that is, $\Vert \triangledown F_{k}(\globalw^{t}_{k}) \Vert^{2} \leq G^{2} + \Vert \triangledown F(\globalw^{t}) \Vert^{2}$ for $k = 0, 1,...m-1$, and $t = 0,..T-1$.
\end{assumption}

In accordance with Theorem 1 in \citep{liconvergence} and our scenario in which all local data stored on clients are used for training in every iteration, we can derive the following convergence result.

\begin{theorem}
    \label{thm:converge}
    Let Assumptions~\ref{assp:lsmooth} to~\ref{assp:gradbound}  hold and $L,\mu, G$ be defined therein. $E$ is the local training epoch. Choose $\gamma=max\{8\frac{L}{\mu},E\}$, and $\eta_{t}=\frac{2}{\mu(\gamma + t)}$. Let $F^{*}$ and $F_{k}^{*}$ be the minimum value of global model $F$ and each local model $F_{k}$ respectively, then:

    \begin{align*}
        & \mathop{\mathbb{E}}[F(\globalw^{T})] - F^{*} 
        \leq  {\frac{2}{\mu^{2}}} \cdot \frac{L}{\gamma +T}(B+D + \frac{\mu^{2}}{4} \mathbb{E}\Vert \globalw^{1}-\globalw^{*}\Vert^{2}).
    \end{align*}
\end{theorem}

where $T$ is the total number of iterations. The expression $B=8 (E-1)^{2} G^{2} + 6L\Gamma$ indicates the combined influence of stochastic optimization and data heterogeneity, since $\Gamma = F^{*} - \sum_{k=1}^{m}{p_{k}F_{k}^{*}}$ effectively quantifies the degree of data heterogeneity. The term $D = 4 \frac{M^2-m^2}{m^2 M^2(M^2-1)} (E-1)^2 G^{2}$ characterizes the impact of random client selection, and $m$ is the number of selected clients, $M$ is the number of total clients.
%
The convergence speed is $O(\frac{1}{T})$. The proof for the theorem is available in Appendix~\ref{convergence_partial}. 
\begin{remark}
    In order to ensure that the upper bound is less than a predefined value $\epsilon$, given as ${\frac{2}{\mu^{2}}} \cdot \frac{L (B+D + \frac{\mu^{2}}{4} \mathbb{E}\Vert \globalw^{1}-\globalw^{*}\Vert^{2})}{\gamma +T} \leq \epsilon$, then the minimal required communication round $T$ must satisfy the condition $T \geq {\frac{2L (B+D + \frac{\mu^{2}}{4} \mathbb{E}\Vert \globalw^{1}-\globalw^{*}\Vert^{2})}{\mu^{2}\epsilon}} - \gamma$.
\end{remark}

\begin{remark}
    With full client participation ($m=M$), the term $D = 4 \frac{M^2-m^2}{m^2 M^2(M^2-1)} (E-1)^2 G^{2}$ in Theorem~\ref{thm:converge} becomes 0. It means full client participation leads to a tighter convergence bound.
\end{remark}

\begin{remark}
    In the absence of our formulation, which implies no additional latent clients, the term $\frac{M^2-m^2}{m^2M^2(M^2-1)}$ in Theorem~\ref{thm:converge} becomes $\frac{M-m}{m(M-1)}$. As $m>1, M>1$, $\frac{M^2-m^2}{m^2M^2(M^2-1)} < \frac{M-m}{m(M-1)}$. It means our approach leads to a faster convergence speed. The analysis is available in Appendix~\ref{convergence_partial}. 
\end{remark}

\begin{table*}[t]
\caption{The test accuracy for FL. Incorporating filter decomposition leads to higher accuracy.}
\centering
  \scalebox{0.84}{
  \label{tab:central}
  \centering
   \begin{tabular}{l|cccccc}
\toprule
                           & \multicolumn{2}{c}{CIFAR10}    & \multicolumn{2}{c}{CIFAR100}    & \multicolumn{2}{c}{Tiny-ImageNet} \\ \cline{2-7} 
                           & (100,2)       & (100,5)        & (100,5)        & (100,20)       & (20,20)       & (20,50)       \\ \hline
FedAvg~\citep{mcmahan2017communication}               & 75.57         & 79.44          & 39.72          &42.14           & 33.47           & 38.67            \\
\rowcolor{Gray} + Decomposition                  & 77.67         & 81.05          & 42.54          & 44.47          & 34.91	          & 41.05         \\ \midrule
FedProx~\citep{li2020federated2}              & 75.83         & 79.35          & 39.74          & 41.97          & 34.29           & 39.81          \\ 
\rowcolor{Gray} + Decomposition                  & 76.98         & 80.93          & 41.77          & 44.13          & 36.17           & 42           \\ \midrule
Scaffold~\citep{karimireddy2020scaffold}                     & 74.96         & 81.74          & 42.48          & 47.49          & 37.27           & 42.98            \\
\rowcolor{Gray} + Decomposition                  & 77.11         & 83.3          & 45.31          & 50.83          & 39.64           & 44.22           \\ \midrule
FedDyn~\citep{acarfederated}                     & 77.04         & 81.81         & 43.48           & 48.8           & 36.77            & 41.98            \\ 
\rowcolor{Gray} + Decomposition                  & 77.88         & 82.82          & 43.57          & 51.05          & 36.7           & 43.56           \\\midrule
FedDC~\citep{gao2022feddc}                    & 77.2         & 82.49          & 43.95           & 50.54           & 36.1            & 41.53            \\
\rowcolor{Gray} + Decomposition      & 77.89         &  83.45   & 43.9          & 52.32          & 37.74           & 43.2           \\ \bottomrule
\end{tabular}
 }
\end{table*}

\section{Experiments}
\label{experiment}
In this section, we demonstrate the efficacy of our proposed approach on three widely used image datasets, namely CIFAR-10, CIFAR-100~\citep{cifar10}, and Tiny-ImageNet~\citep{le2015tiny}. Our approach outperforms baseline methods in terms of global test accuracy, both for IID and non-IID cases. 
We also explore the potential of our approach in personalized FL with local filter atoms.

\subsection{Experimental Setup}
\paragraph{Datasets and Models.} We conduct a series of experiments on three image datasets: CIFAR-10, CIFAR-100~\citep{cifar10}, and Tiny-Imagenet~\citep{le2015tiny}. We utilize LeNet~\cite{mcmahan2017communication} for CIFAR10 and CIFAR100, pre-trained ResNet18~\cite{he2016deep} for Tiny-ImageNet. In each experiment, we have set the number of filter atoms $m_a$ in our approach to 9.
Details of the datasets and models are presented in Appendix~\ref{exp:setting}.

\paragraph{Data Partitions.}
In the FL setting, we assume the existence of a central server and a set of $M=100$ local clients for CIFAR-10/100 and $M=20$ local clients for Tiny-ImageNet, each client holding a subset of the total dataset. In each communication round, a random subset of $10\%$ clients is selected for local training, and we set the number of local epochs $E=5$. 
The communication round for CIFAR-10/100 and Tiny-ImageNet are $T=600$ and $T=10$.
In the non-IID case, we follow~\citep{collins2021exploiting, mcmahan2017communication} to partition the data in the following manner: for CIFAR-10, the data are divided into 100 clients with either 2 or 5 classes on each client, denoted as $(100, 2)$ and $(100, 5)$, respectively.
Similarly, the data of CIFAR-100 are split into $(100, 5)$, $(100, 20)$ and the data of Tiny-ImageNet are divided into $(20, 20)$ and $(20, 50)$, respectively. 
This partitioning method introduces more extreme data heterogeneity. 
We also adopt the data partition method with Dirichlet distribution~\citep{yurochkin2019bayesian} and provide the results in the Appendix~\ref{sec:extra_exp}.

\paragraph{Compared Methods.} For the evaluation of the global model, we apply the proposed filter decomposition method on five approaches including FedAvg~\citep{mcmahan2017communication}, FedProx~\citep{li2020federated2}, FedDyn~\citep{acarfederated}, FedDC~\citep{gao2022feddc}, and Scaffold~\citep{karimireddy2020scaffold}. The results of more methods are in the Appendix~\ref{sec:extra_exp}.
To evaluate the performance of personalized models, we compare our approach with several baseline methods. These baselines include Local-only, where each client is trained independently without any communication, as well as global models fine-tuned on local data, such as FedAvg+FT and FedProx+FT. We also consider the multi-task method Ditto~\citep{li2021ditto} and parameter decoupling approaches, including LG-FedAvg~\citep{lgfedavg}, FedPer~\citep{fedper}, FedRep~\citep{collins2021exploiting}, and FedPAC~\citep{fedpac}.

\subsection{Accuracy Comparison}
The comparison between models with and without filter decomposition is shown in Table~\ref{tab:central}. 
The test accuracy is computed on the whole test set, which includes all classes. All experiments are conducted with three random seeds, and the results presented in the table are the mean values. Standard deviations are not included in the table as they are all relatively small. This experiment demonstrates the effectiveness of filter decomposition when applying our method to five baseline methods~\cite{acarfederated, gao2022feddc, karimireddy2020scaffold, li2020federated2, mcmahan2017communication}. Despite each of these methods having its distinct local objective, the straightforward application of filter decomposition results in significant accuracy improvements.

\begin{figure}[t]
  \centering
 \begin{subfigure}[b]{0.23\textwidth}
     \centering
     \includegraphics[trim={0pt 0pt 0pt 0pt}, clip, width=\textwidth]{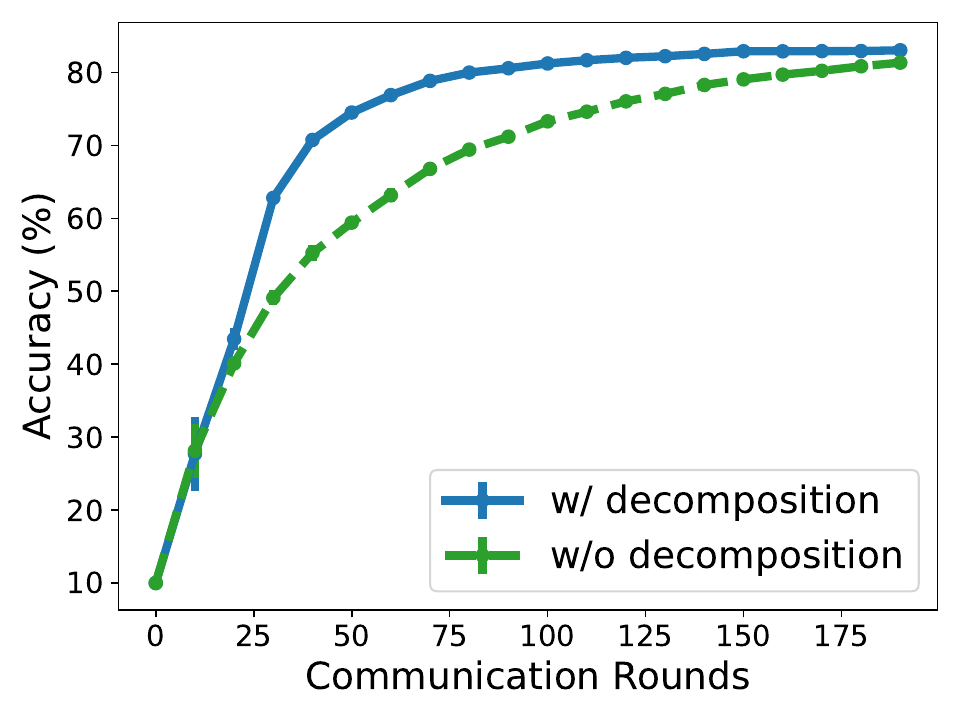}
     \caption{}
 \end{subfigure}
 \begin{subfigure}[b]{0.23\textwidth}
     \centering
     \includegraphics[trim={0pt 0pt 0pt 0pt}, clip, width=\textwidth]{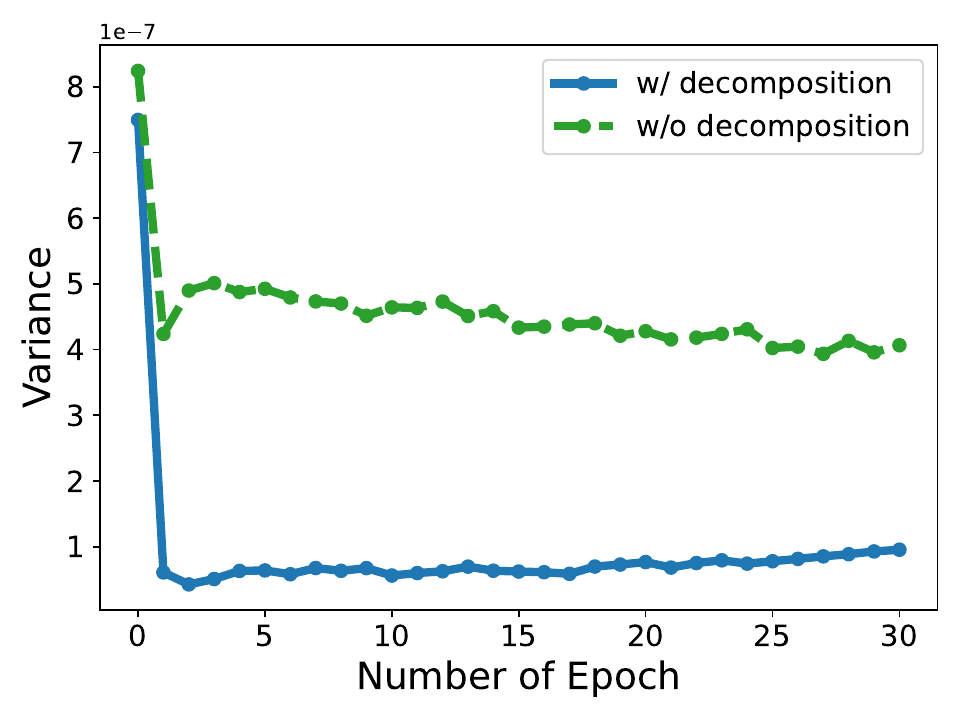}
     \caption{}
 \end{subfigure}
 \begin{subfigure}[b]{0.23\textwidth}
     \centering
     \includegraphics[trim={0pt 0pt 0pt 0pt}, clip, width=\textwidth]{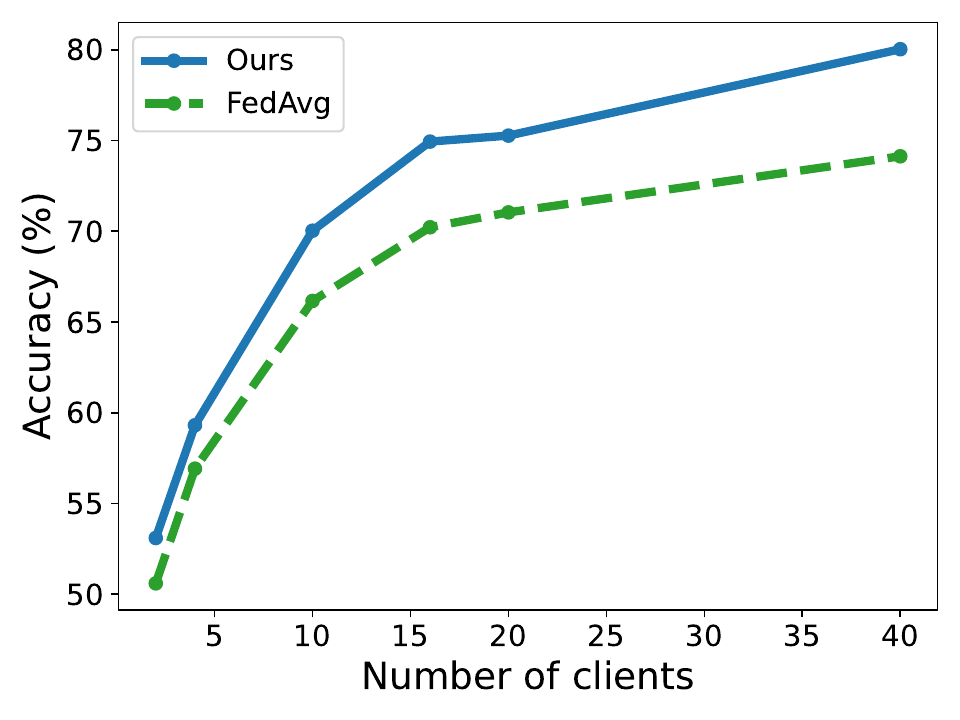}
     \caption{}
 \end{subfigure}
 \begin{subfigure}[b]{0.23\textwidth}
     \centering
     \includegraphics[trim={0pt 0pt 0pt 0pt}, clip, width=\textwidth]{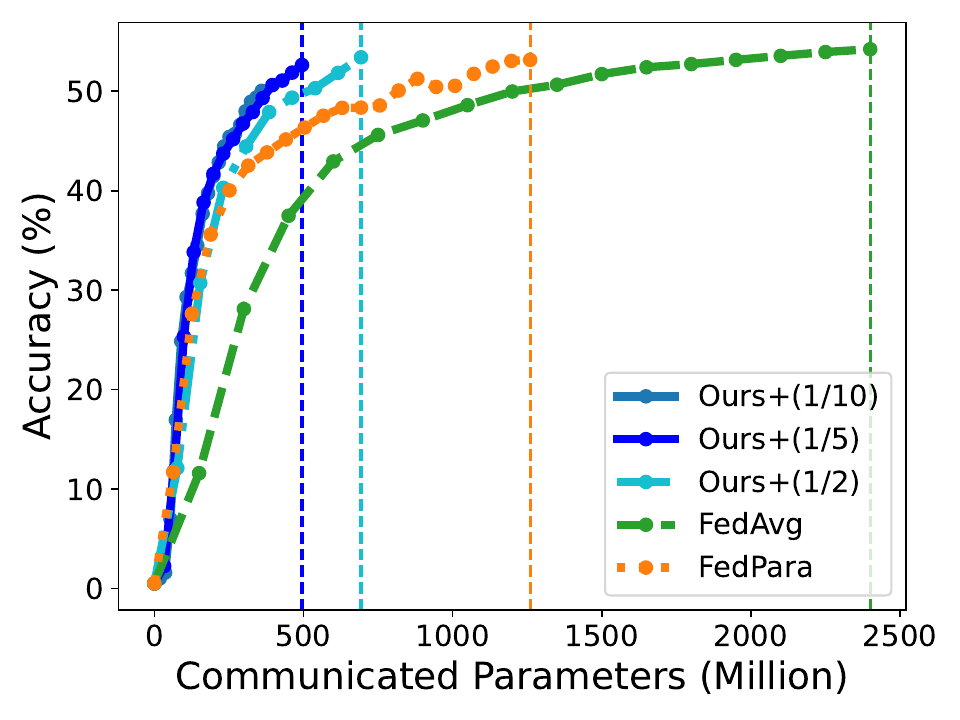}
     \caption{}
 \end{subfigure}
  \caption{Filter decomposition naturally introduces extra latent clients, offering several advantages to FL: (a) It accelerates the model's convergence speed and increases its accuracy. 
  (b) Employing filter decomposition minimizes variance and maintains this reduction. (c) As the number of clients increases, the test accuracy improves. With filter decomposition, additional latent clients boost this accuracy even further. (d) Our approach also reduces communication costs, as evidenced by a comparison of the parameters communicated to reach the same accuracy.}
  \label{fig:save_bandwidth}
\end{figure}

We plot the test accuracy with respect to the number of epochs for FedDyn using CIFAR10 (100, 5) as a demonstration, as shown in Figure~\ref{fig:save_bandwidth}(a). Filter decomposition enables FedDyn to attain the same level of accuracy with reduced communication. This phenomenon can be attributed to faster convergence resulting from the presence of additional latent clients, consequently leading to reduced variance. We offer a more detailed discussion in Section~\ref{sec:more_clients}.  

\subsection{Application in Personalized FL}
We demonstrate the efficacy of our approach in the personalized federated learning setting. Each client maintains both local filter atoms $\atom_{l}$ and global filter atoms $\atom_{g}$. We evaluate the accuracy of local models constructed with preserved local filter atoms and shared global atom coefficients. 
For the baseline methods FedAvg and FedProx, we fine-tune the models on clients' data for 10 epochs. 

The experimental results are presented in Table~\ref{tab:personalization}, highlighting the best accuracy in bold and the second-best accuracy with underlines. In the majority of cases, our approach exhibits higher accuracy compared to other methods. Particularly, in challenging tasks such as Tiny-ImageNet, our method outperforms the baselines by a large margin. 
The experiments demonstrate that the filter atoms can effectively capture personalized local knowledge and are substantial for model personalization.

\subsection{Communication Efficiency}
\label{sec:comm_eff}
Our approach effectively lowers communication costs by transmitting atom coefficients less frequently compared to filter atoms.
In this experiment, employing CIFAR100 (100,20) task, our approach communicates filter atoms every round but transmits whole models every 2nd, 5th, or 10th round, represented by $\beta=1/2$, $\beta=1/5$ and $\beta=1/10$, respectively. Figure~\ref{fig:save_bandwidth}(d) displays the number of parameters communicated to attain the same accuracy across various methods, where the x-axis is the number of transmitted parameters. Our approach with $\beta=1/5$ communicates only about $500$ million parameters to achieve an accuracy of $52\%$, while FedAvg requires $2,500$ million parameters, which is five times more than our proposed method. FedPara~\cite{hyeonfedpara} communicates $1,250$ parameters to achieve the same accuracy, which is over twice the number of parameters than our method.

\begin{table*}
\caption{The test accuracy for personalized FL.}
\centering
  \scalebox{0.84}{
  \label{tab:personalization}
  \centering
   \begin{tabular}{l|cccccc}
\toprule
                           & \multicolumn{2}{c}{CIFAR10}    & \multicolumn{2}{c}{CIFAR100}    & \multicolumn{2}{c}{Tiny-ImageNet} \\ \cline{2-7} 
                           & (100,2)       & (100,5)        & (100,5)        & (100,20)       & (1000,20)       & (1000,50)       \\ \hline
Local                      & 88.00         & 76.00          & 76.10           & 41.90           & 12.89           & 6.13            \\
FedAvg~\citep{mcmahan2017communication} + FT                & 93.53         & \underline{86.85}          & \textbf{85.29}          & \textbf{64.10}           & \underline{30.77}           & 14.6            \\
FedProx~\citep{li2020federated2} + FT               & 92.46         & 85.31          & 80.28          & 59.07          & 28.69           & \underline{14.96}           \\ \midrule
Ditto~\citep{li2021ditto}                      & \underline{93.67}         & 85.75          & 80.37          & 62.56          & 21.93           & 9.64            \\
FedPer~\citep{fedper}                     & 92.60         & 83.30          & 76.00           & 37.70           & 14.20            & 6.15            \\
FedRep~\citep{collins2021exploiting}                     & 89.34         & 79.25          & 78.53           & 57.89           & 13.90            & 6.14            \\
FedPAC~\citep{fedpac}                     & 91.20         & 85.30          & 80.90           & 61.64           & 13.90         & 6.15         \\
LG-FedAvg~\citep{lgfedavg}                  & 88.50         & 72.80          & 73.00           & 41.10           & 12.44           & 5.99            \\ \midrule
Ours     & \textbf{94.14}         & \textbf{87.49} & \underline{84.28}          & \underline{63.28}          & \textbf{41.57}           & \textbf{24.32}           \\ \bottomrule
\end{tabular}
 }
\end{table*}

\section{Discussion}
In this section, we provide more evidence of how the proposed method reduces variance and leads to faster convergence. We also discuss the impact of hyper-parameters on our approach and provide an intuitive explanation of its effectiveness by examining the loss landscape.

\begin{figure*}[t]
  \centering
   \begin{subfigure}[b]{0.28\textwidth}
    \centering
    \includegraphics[trim={0pt 0pt 0pt 0pt}, clip, width=\textwidth]{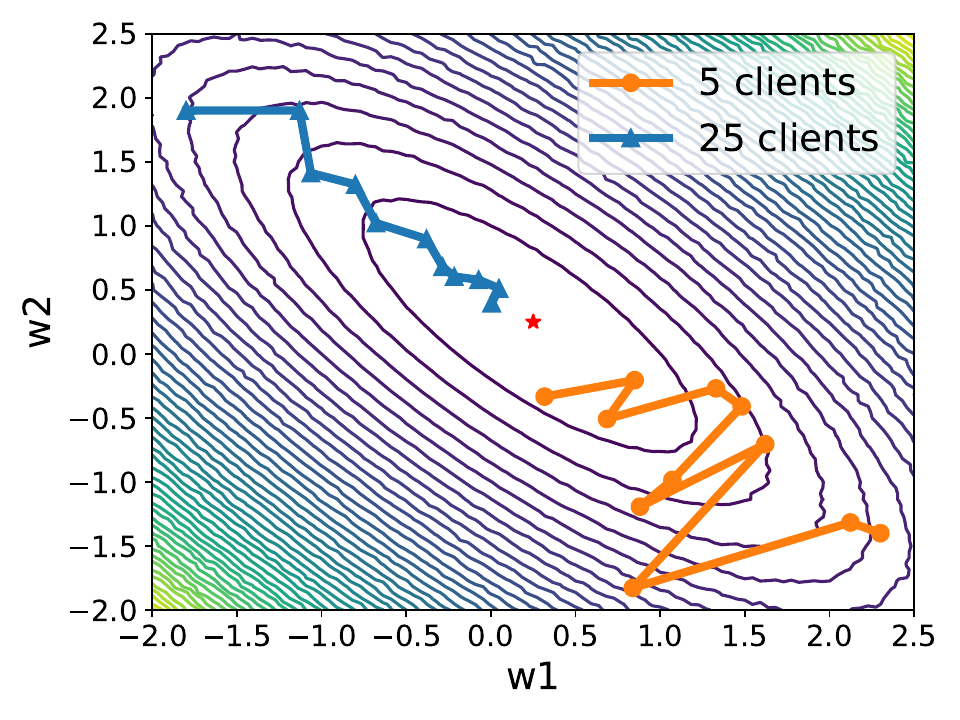} 
    \caption{Fast Convergence}
  \end{subfigure}
  \begin{subfigure}[b]{0.28\textwidth}
    \centering
    \includegraphics[trim={0pt 0pt 0pt 0pt}, clip, width=\textwidth]{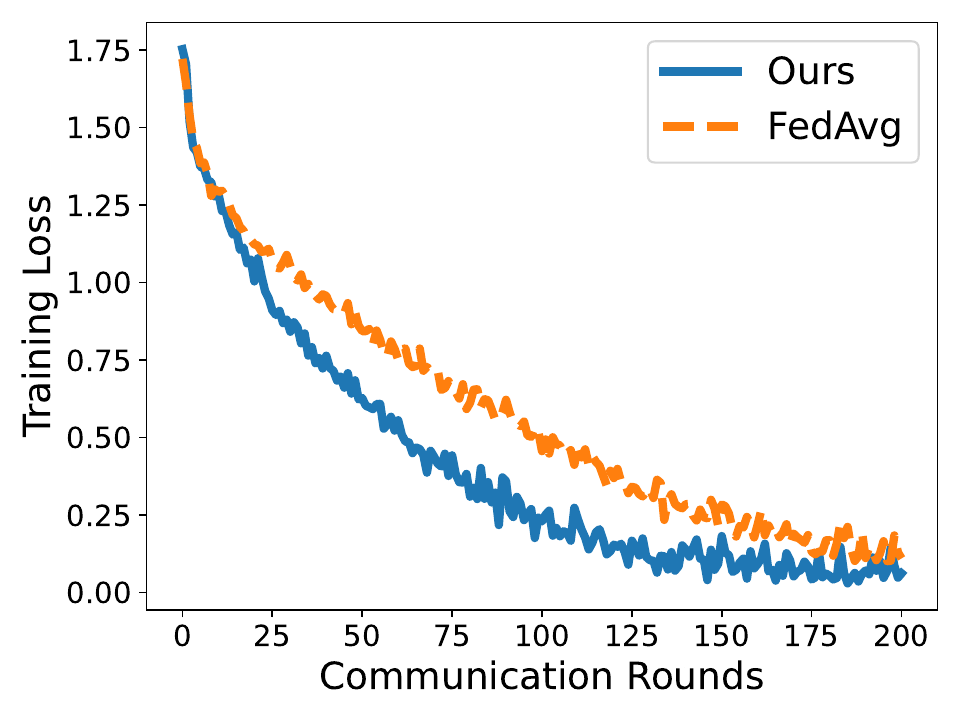} 
    \caption{Training loss}
  \end{subfigure}
  \begin{subfigure}[b]{0.28\textwidth}
    \centering
    \includegraphics[trim={0pt 160pt 0pt 160pt}, clip, width=\textwidth]{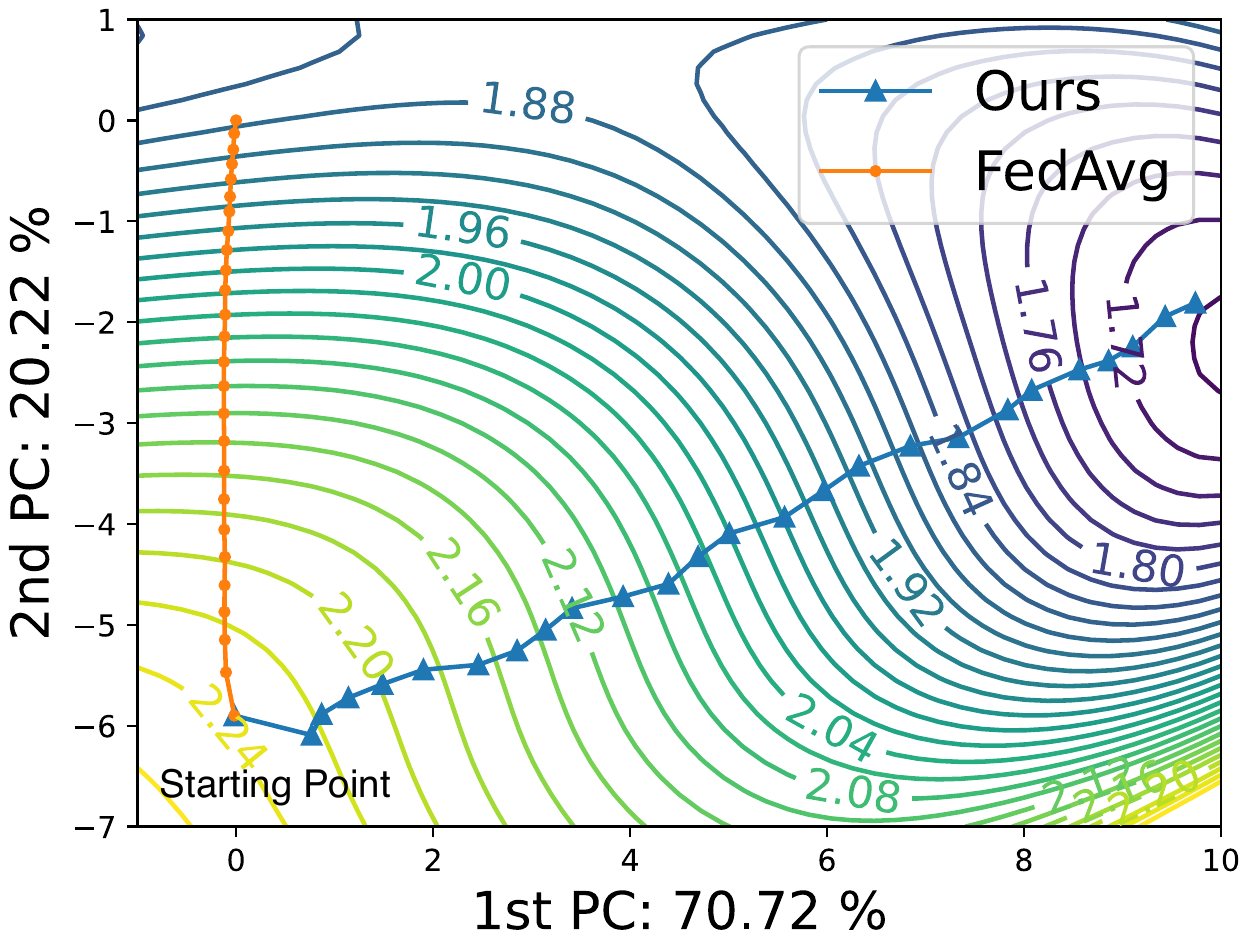} 
    \caption{Loss landscape}
  \end{subfigure}
  \caption{(a) The loss landscape shows that additional clients result in reduced variance and enhanced training stability, resulting in faster convergence.
  We employ FedAvg as an example to depict the impact of filter decomposition on the training loss. (b) Our filter decomposition method shows lower training loss than FedAvg. (c) As visualized in the loss landscape, our method achieves lower loss faster than FedAvg. 
  }
  \label{fig:loss_landscape}
\vspace{-8pt}
\end{figure*}

\subsection{Influence of Additional Clients}
\label{sec:more_clients}

We provide additional results to demonstrate that a greater number of involved clients reduces aggregation variance and consequently improves global test accuracy. 

\paragraph{Validation of reduced variance.}
During each communication round, various factors, including stochastic optimization, random client selections, and data shuffling, can all contribute to the randomness in the aggregated models. According to Section~\ref{sec:reduce_var}, for the model without filter decomposition, the aggregated parameter is $\param_{\extractor}$, and the aggregated parameter of the model with filter decomposition is $\param_{\extractor}'$. In this experiment, we estimate the variance of the aggregated parameter $\mathbb{E} ||\param_{\extractor} - \mathbb{E} (\param_{\extractor})||^2$ by $\frac{1}{N_c} \sum_{i=1}^{N_c}||\param_{\extractor}^{i} - \frac{1}{N_c} \sum_{i=1}^{N_c} \param_{\extractor}^{i}||^2$, where $N_c$ is the number of repeated local training time. We chose to repeat the local training 10 times, resulting in the generation of 10 distinct aggregated models $\param_{\extractor}$ and $\param_{\extractor}'$. The comparison of variance between models with and without filter decomposition is shown in Figure~\ref{fig:save_bandwidth}(b). The model with filter decomposition reaches a small variance after just one communication round, and this low variance is consistently maintained.


\paragraph{Synthetic Experiment.} We design a synthetic experiment to validate that more clients lead to a decrease in the variance of the aggregated model. This is a classification task with positive samples from $\mathcal{N}(\mu_+, \Sigma)$ and negative samples from $\mathcal{N}(\mu_-, \Sigma)$, where $\mu_+=\begin{bmatrix} 2 & 2\end{bmatrix}$ and $\mu_-=\begin{bmatrix} -2 & -2\end{bmatrix}$. The model is one fully connected layer with two parameters so that we can better visualize the result in a 2D plot. And the loss function is the mean square error. The experiment is conducted with 5 models and 25 models, and Figure~\ref{fig:loss_landscape}(a) displays the loss landscape of the averaged model. As the number of clients increases, the aggregated model becomes more robust and converges toward the optimal point within fewer training rounds.

\paragraph{More involved models result in higher accuracy.}
In this experiment, each client contains 5 classes of CIFAR-10 data and we report the final result at the $200$th round. The number of the involved clients increases from 2 to 40, We observe in Figure~\ref{fig:save_bandwidth}(c) that the accuracy improves as the number of involved clients increases, even without filter decomposition. 
According to $\param_{\extractor}'=\sum_{k=1}^{m} \frac{n_k^2}{n^2} \param_{k, \extractor} + \sum_{k_1 \neq k_2}^{m} \frac{n_{k_1} \cdot n_{k_2}}{n^2} \param_{k_1, k_2, \extractor}$, with filter decomposition, $m$ aggregated models provide $m^2-m$ additional latent clients. This explains why models with filter decomposition experience greater accuracy improvement as the number of participating clients increases.

\subsection{Influence of different $m_a$}
Adjusting $m_a$ is a trade-off between model accuracy and training parameters. We investigate the influence of the number of filter atoms $m_a$ on the model accuracy within the pFL framework. 
Larger values of $m_a$ correspond to higher accuracy. Smaller values of $m_a$ lead to fewer involved parameters, thus less computational resource required for training and less communication overhead. We provide details in Appendix~\ref{sec:extra_exp}.



\subsection{Training Loss}
We compare the training loss of FedAvg with our filter decomposition method, empirically validating that the proposed approach leads to an increase in client diversity and exhibits faster convergence speed.
The training loss of both FedAvg and our method is depicted in Figure~\ref{fig:loss_landscape}(b). Our approach exhibits lower training loss than FedAvg starting from the 20th round.

To gain further insights into the faster convergence of our approach, we examine the loss landscape~\citep{li2018visualizing} by plotting the training trajectories of FedAvg and our approach for the first 25 epochs, as depicted in Figure~\ref{fig:loss_landscape}(c). The contour map represents the loss reduction from 2.24 (bottom left) to 1.72 (top right). 
Our approach demonstrates faster movement towards a lower loss compared to FedAvg within the same number of training rounds.
\section{Conclusion}
This paper tackled the data heterogeneity challenges in FL. Different from conventional FL aggregation methods, our approach utilizes decomposed filters, consisting of filter atoms and atom coefficients, to reconstruct a global model through aggregated atoms and coefficients. This reconstructed global model effectively reduces the variance of the global model by introducing additional model variants, thereby providing a faster convergence. Through extensive experiments conducted on CIFAR-10, CIFAR-100, and Tiny-ImageNet datasets, we have demonstrated that our filter decomposition method can easily integrate with various FL approaches, leading to improved accuracy. These results highlight the effectiveness and superiority of our approach in the context of federated learning.

\paragraph{Limitations.} Our approach is grounded in the use of CNN. While our method can be extended to linear layers and ViT, such applications would require further validation not covered in our current study.




\section*{Impact Statement}
This paper presents work whose goal is to advance the field of 
Machine Learning. There are many potential societal consequences 
of our work, none which we feel must be specifically highlighted here.

\nocite{langley00}

\bibliography{example_paper}
\bibliographystyle{icml2025}

\newpage
\appendix
\onecolumn

\section{Overview of Supplementary}
The appendix provides supplementary to the main content and includes the following details:
\begin{itemize}
    \item Algorithm design is described in Section~\ref{appx:algorithm}.
    \item The analysis of reduced variance caused by filter decomposition is presented in Section~\ref{theory:variance}.
    \item Convergence analysis is discussed in detail in Sections~\ref{theory:convergence}, \ref{convergence_partial}.
    \item The application of our method to further reduce communication time is explored in Section~\ref{sec:fast_slow}.
    \item Additional experimental details can be found in Section~\ref{exp:setting}.
    \item Further experimental results are provided in Section~\ref{sec:extra_exp}.
\end{itemize}

\section{Algorithm Design}
\label{appx:algorithm}
During each communication round, the model separately trains and aggregates the $\coeff$, $\atom$, and $\param_h$. The global convolutional filter is then formed by multiplying $\coeff^{t+1}$ and $\atom^{t+1}$ of selected $m$ clients. 

The main training procedure is summarized in Algorithm~\ref{alg:algorithm}. $M$ is the total number of clients, and $C$ is the ratio of selected clients at each communication round. $T$ is the total communication round. $\globalw^{0}$ are the initialized parameters, and $\eta_0$ is the initial learning rate. $B, E$ are the local training batch size and local training epoch.

\begin{algorithm}[]
\caption{Global Communicated Filter Atoms and Atom Coefficients}
\label{alg:algorithm}
\textbf{Input}: $M, T, \globalw^{0}, \eta_0, B, C, E$ \\
\textbf{Server:}
\begin{algorithmic} 
\FOR{$t = 0, 1, 2, \dots, T-1$}
\item $m \xleftarrow{} max(C\cdot M,1)$
\item $S_{t} \xleftarrow{} $Random set of $m$ clients
\FOR{k $\in S_{t}$}
\item $\coeff^{t+1}_{k}, \atom^{t+1}_{k}, \param_{k, h}^{t+1} \xleftarrow{} Client(k, \coeff^{t}, \atom^{t}, \param_{h}^{t})$
\ENDFOR
\STATE{\emph{// weight aggregation}}
\item $\coeff^{t+1} \xleftarrow{} \sum_{k=1}^{m} \frac{n_k}{n} \coeff^{t+1}_k $
\item $\atom^{t+1} \xleftarrow{} \sum_{k=1}^{m} \frac{n_k}{n} \atom^{t+1}_k $
\item $\param_{h}^{t+1} \xleftarrow{} \sum_{k=1}^{m} \frac{n_k}{n}  \param_{k, h}^{t+1} $

\STATE{\emph{// re-construct global model}}
\item $\param_{\extractor}^{t+1} \xleftarrow{} \coeff^{t+1} \times \atom^{t+1}$
\ENDFOR\\

\end{algorithmic}
\textbf{Client:}
\begin{algorithmic}
\FOR{$i=0,1,2,\dots, E-1$}
\FOR{$b \in B$}
\item $\coeff^{t+1}_{k} \xleftarrow{} \coeff^{t}_k - \eta_t \nabla_{\coeff^{t}_k} F_k$
\item $\atom^{t+1}_{k} \xleftarrow{} \atom^{t}_k - \eta_t \nabla_{\atom^{t}_k} F_k$
\item $\param_{k, h}^{t+1} \xleftarrow{} \param_{k, h}^{t} - \eta_t \nabla_{\param_{k, h}^{t}} F_k$
\ENDFOR
\ENDFOR
\STATE \textbf{return} $\coeff^{t+1}_{k}, \atom^{t+1}_{k}, \param_{k, h}^{t+1}$
\end{algorithmic}
\end{algorithm}

The reconstruction of convolutional filters through the multiplication of $\coeff$ and $\atom$ does not introduce any additional computational overhead. The model represents the convolutional layer as a combination of one atom coefficient layer and one filter atom layer. These two components are multiplied together seamlessly during the forward pass of the model.

\section{Proof of Reduced Global Model Variance}
\label{theory:variance}
According to (\ref{eq:dcf_weight_update}), the reconstruction of decomposed filters results in a natural incorporation of additional local model variants, which is
\vspace{-3pt}
\begin{align*}
    \param_{\extractor}^{t+1} & = (\sum_{k=1}^{m} \frac{n_k}{n} \coeff^{t+1}_k) \times (\sum_{k=1}^{m} \frac{n_k}{n} \atom^{t+1}_k) \nonumber \\
            & = \sum_{k=1}^{m} \frac{n_k^2}{n^2} \param_{k, \extractor}^{t+1} + \sum_{k_1 \neq k_2}^{m} \frac{n_{k_1} \cdot n_{k_2}}{n^2} \param_{k_1, k_2, \extractor}^{t+1},
\end{align*} 
where $\param_{k, \extractor}^{t+1}=\coeff^{t+1}_k \times \atom^{t+1}_k$ and $\param_{k_1, k_2, \extractor}^{t+1}=\coeff^{t+1}_{k_1} \times \atom^{t+1}_{k_2}$.
Compared with the weight update of FedAvg, this formulation contains both averaging of selected clients represented in the first term and extra reconstructed latent clients in the second term. 

Denote the parameter obtained by FedAvg as $\param_{\extractor}=\sum_{k=1}^{m} \frac{n_k}{n} \param_{k, \extractor}=\sum_{k=1}^{m} p_k \param_{k, \extractor}$, and parameter obtained by (\ref{eq:dcf_weight_update}) as $\param_{\extractor}'=\sum_{k=1}^{m} \frac{n_k^2}{n^2} \param_{k, \extractor} + \sum_{k_1 \neq k_2}^{m} \frac{n_{k_1} \cdot n_{k_2}}{n^2} \param_{k_1, k_2, \extractor}=\sum_{k=1}^{m} p_k^2 \param_{k, \extractor} + \sum_{k_1 \neq k_2}^{m} p_{k_1}p_{k_2} \param_{k_1, k_2, \extractor}=\sum_{k=1}^{m} p_k^2 \param_{k, \extractor} + \sum_{k'=1}^{m^2-m} p'_{k'} \param_{k', \extractor}$, where $\{p'_{k'}\}_{k'=1}^{m^2-m}=\{p_{k_1}p_{k_2}\}_{k_1 \neq k_2}^{m}$.

We consider $\param_{k, \extractor}$ and $\param_{k_1, k_2, \extractor}$ as independent random variables. They also share the same expectation denoted as $\hat{\param}_{\extractor}=\mathbb{E} (\param_{k, \extractor})$. By the definition of variance, we have,

\begin{align*}
    & \mathbb{E} ||\param_{\extractor}' - \mathbb{E} (\param_{\extractor}')||^2 \\
    = & \mathbb{E} ||\param_{\extractor}'||^2 - ||\mathbb{E} (\param_{\extractor}')||^2  \\
    = & \mathbb{E} ||\sum_{k=1}^{m} p_k \param_{k, \extractor}'||^2 - ||\hat{\param}_{\extractor}||^2  \\
    = & \mathbb{E} ||\sum_{k=1}^{m} p_k^2 \param_{k, \extractor} + \sum_{k'=1}^{m^2-m} p'_{k'} \param_{k', \extractor}||^2 - ||\hat{\param}_{\extractor}||^2  \\
    = & (\sum_{k=1}^{m} p_k^4 + \sum_{k'=1}^{m^2-m} p^{'2}_{k'}) \mathbb{E}||\param_{\extractor}||^2 \\
    & + (\sum_{k_1\neq k_2}^{m} p_{k_1}^2p_{k_2}^2 + \sum_{k'_1 \neq k'_2}^{m^2-m} p'_{k'_1}p'_{k'_2} \\
    & + 2(\sum_{k=1}^{m} p_k^2)(\sum_{k'=1}^{m^2-m}  p^{'}_{k'})) ||\hat{\param}_{\extractor}||^2  - ||\hat{\param}_{\extractor}||^2 \\
    = & (\sum_{k=1}^{m} p_k^2)^2 \mathbb{E}||\param_{\extractor}||^2 + ((\sum_{k'=1}^{m^2-m} p'_{k'})^2 \\
    & + 2(\sum_{k=1}^{m} p_k^2)(\sum_{k'=1}^{m^2-m}  p^{'}_{k'})) ||\hat{\param}_{\extractor}||^2  \\
    & - ||\hat{\param}_{\extractor}||^2 \\
    = & (\sum_{k=1}^{m} p_k^2)^2 \mathbb{E}||\param_{\extractor}||^2 + ((\sum_{k=1}^{m} p_k)^2 \\
    & - \sum_{k=1}^{m} p_k^2)((\sum_{k=1}^{m} p_k)^2 + \sum_{k=1}^{m} p_k^2)||\hat{\param}_{\extractor}||^2  \\
    & - ||\hat{\param}_{\extractor}||^2,
\end{align*} 
where $||\hat{\param}_{\extractor}||^2=||\mathbb{E} (\param_{\extractor})||^2$. Similarly,

\begin{align*}
    & \mathbb{E} ||\param_{\extractor} - \mathbb{E} (\param_{\extractor})||^2 \\
    & = \mathbb{E} ||\param_{\extractor}||^2 - ||\mathbb{E} (\param_{\extractor})||^2  \\
    & = \mathbb{E} ||\sum_{k=1}^{m} p_k \param_{k, \extractor}||^2 - ||\hat{\param}_{\extractor}||^2  \\
    & = \mathbb{E} \left[\sum_{k=1}^{m} p_k^2 ||\param_{k, \extractor}||^2 + \sum_{k_1\neq k_2}^{m} p_{k_1}p_{k_2} \param_{k_1, \extractor} \cdot \param_{k_2, \extractor}\right ] - ||\hat{\param}_{\extractor}||^2  \\
    & = \sum_{k=1}^{m} p_k^2 \mathbb{E} ||\param_{\extractor}||^2 + \sum_{k_1\neq k_2}^{m} p_{k_1}p_{k_2} \mathbb{E} \left[\param_{k_1, \extractor}\right] \cdot \mathbb{E} \left[\param_{k_2, \extractor}\right] - ||\hat{\param}_{\extractor}||^2  \\
    & = \sum_{k=1}^{m} p_k^2 \mathbb{E} ||\param_{\extractor}||^2 + ((\sum_{k=1}^{m} p_k)^2 - \sum_{k=1}^{m} p_k^2) ||\hat{\param}_{\extractor}||^2 - ||\hat{\param}_{\extractor}||^2.
\end{align*} 

Then, combine above two equations and with $\sum_{k=1}^{m} p_k=1$, we have 
\begin{align*}
     & {\scriptsize \mathbb{E} ||\param_{\extractor}' - \mathbb{E} (\param_{\extractor}')||^2 - \mathbb{E} ||\param_{\extractor} - \mathbb{E} (\param_{\extractor})||^2} \\
    =  & {\scriptsize \left [ (\sum_{k=1}^{m} p_k^2)^2 \mathbb{E}||\param_{\extractor}||^2 
    + ((\sum_{k=1}^{m} p_k)^2 - \sum_{k=1}^{m} p_k^2)((\sum_{k=1}^{m} p_k)^2 
    + \sum_{k=1}^{m} p_k^2)||\hat{\param}_{\extractor}||^2 - ||\hat{\param}_{\extractor}||^2 \right]} \\
     & {\scriptsize - \left [ \sum_{k=1}^{m} p_k^2 \mathbb{E} ||\param_{\extractor}||^2 + ((\sum_{k=1}^{m} p_k)^2 - \sum_{k=1}^{m} p_k^2) ||\hat{\param}_{\extractor}||^2 - ||\hat{\param}_{\extractor}||^2 \right]} \\
    = & {\scriptsize(\sum_{k=1}^{m} p_k^2)(\sum_{k=1}^{m} p_k^2 - 1) \mathbb{E} ||\param_{\extractor}||^2 + ((\sum_{k=1}^{m} p_k)^2 - \sum_{k=1}^{m} p_k^2) (\sum_{k=1}^{m} p_k^2)||\hat{\param}_{\extractor}||^2} \\
    =  & {\scriptsize -(1 - \sum_{k=1}^{m} p_k^2) (\sum_{k=1}^{m} p_k^2) (\mathbb{E} ||\param_{\extractor}||^2 - ||\hat{\param}_{\extractor}||^2)} \\
    \leq & 0,
\end{align*}

where the last inequality is because of the non-negativity of the variance and $0<\sum_{k=1}^{m} p_k^2 \leq \left(\sum_{k=1}^{m} p_k \right)^2 = 1$. Thus,
\begin{align*}
    \mathbb{E} ||\param_{\extractor}' - \mathbb{E} (\param_{\extractor}')||^2 
    \leq & \mathbb{E} ||\param_{\extractor} - \mathbb{E} (\param_{\extractor})||^2.
\end{align*}

It verifies that as the number of aggregated clients increases, the variance of the global model decreases. Additionally, our method introduces additional latent clients, which naturally contributes to a reduction in the variance of the global model.

\section{Convergence Analysis}
\label{theory:convergence}
\newcommand{\weight}{\mathbf{w}}
\newcommand{\Bweight}{\Bar{\mathbf{w}}}
\newcommand{\expect}{\sum_{k=1}^{m} p_{k}}

\subsection{Formulation}

Recall the definitions from the main content, the global objective function is defined as
\begin{align*}
    F\left(\globalw \right)=\sum_{k=1}^{m} p_k \cdot F_k(\localw).
\end{align*}

The local objective function $F_k(\cdot)$ is given by
\begin{align*}
F_k(\localw) = \frac{1}{n_k} \sum_{j=1}^{n_k} \mathcal{L}(\coeff_k, \atom_k, \param_{k, h}; \mathbf{x}_j, y_j),
\end{align*}
where $\localw = \{\coeff_k, \atom_k, \param_{k, h}\}$. And the model is updated as
\begin{align*}
    \begin{bmatrix} \coeff^{t+1}_k \\ \atom^{t+1}_k \\ \param_{k, h}^{t+1} \end{bmatrix} \xleftarrow{} 
    \begin{bmatrix}\coeff^{t}_k - \eta_t \nabla_{\coeff^{t}_k} F_k \\ \atom^{t}_k - \eta_t \nabla_{\atom^{t}_k} F_k \\ \param_{k, h}^{t} - \eta_t \nabla_{\param_{k, h}^{t}} F_k \end{bmatrix}.
\end{align*}

The model separately aggregates the $\coeff$, $\atom$, and $\param_h$,
\begin{align*}
    \begin{bmatrix} 
        \coeff^{t+1} \\ 
        \atom^{t+1} \\ 
        \param_{h}^{t+1} 
    \end{bmatrix} 
    \xleftarrow{} 
    \begin{bmatrix} 
        \sum_{k=1}^{m} p_k \coeff^{t+1}_k \\ 
        \sum_{k=1}^{m} p_k \atom^{t+1}_k \\ 
        \sum_{k=1}^{m} p_k  \param_{k, h}^{t+1} 
    \end{bmatrix}.
\end{align*}

Thus, we have $\globalw^{t+1}=\{\coeff^{t+1}, \atom^{t+1}, \param_{h}^{t+1}\}$ and $\globalw^{t+1}=\sum_{k=1}^{m} p_k \localw^{t+1}$. For convenience, we define 
$g^{t}=\sum_{k=1}^{m} p_k \nabla F_{k}(\localw^{t})$, where $\nabla F_{k}(\localw^{t})=\{\nabla_{\coeff_k} F_k(\coeff_k), \nabla_{\atom_k} F_k(\atom_k), \nabla_{\param_{k, h}} F_k(\param_{k, h}) \}$.

\subsection{Analysis on consecutive steps}

To bound the expectation of the global objective function at time $T$ from its optimal value, we first consider analyzing the global weight from the optimal weights by calculating single-step SGD:
\begin{align}
\label{lamma1}
    \Vert \globalw^{t+1}- \globalw^{*}\Vert^{2} & = \Vert \globalw^{t}- \globalw^{*} -\eta_{t} g_{t} \Vert^{2} \nonumber \\
    & = \Vert \globalw^{t}- \globalw^{*} \Vert^{2} - 2\eta_{t} \langle \globalw^{t}- \globalw^{*}, g_{t} \rangle + \eta^{2}_{t} \Vert g_{t} \Vert ^{2}.
\end{align}

The second term of (\ref{lamma1}) can be expressed as
\begin{align}
\label{lamma1:A2}
     & {\scriptsize - 2\eta_{t} \langle \globalw^{t}- \globalw^{*}, g_{t} \rangle} \nonumber \\
    = & {\scriptsize-2 \eta_{t} \expect \langle \globalw^{t}- \globalw^{*}, \nabla F_{k} (\localw^{t})\rangle} \nonumber \\
    = &  -2 \eta_{t} \expect \langle \globalw^{t}- \localw^{t}, \nabla F_{k} (\localw^{t}) \rangle \nonumber \\
    & -2 \eta_{t} \expect \langle \localw^{t} - \globalw^{*}, \nabla F_{k} (\localw^{t}) \rangle.
\end{align}


By Cauchy-Schwarz inequality and AM-GM inequality, we have
\begin{equation}
\label{lamma1:A2-1}
\begin{split}
    -2 \langle \globalw^{t}- \localw^{t}, \nabla F_{k} (\localw^{t}) \rangle & \leq \frac{1}{\eta_{t}} \Vert \globalw^{t}- \localw^{t} \Vert^{2} 
         + \eta_{t} \Vert \nabla F_{k} (\localw^{t}) \Vert^{2}.
\end{split}
\end{equation}

By the $\mu$-strong convexity of $F_{k}(\cdot)$, with $v = \globalw^{*}$ and $w = \localw^{t}$, we have
\begin{equation}
\label{lamma1:A2-2}
\begin{split}
    & - \langle \localw^{t} - \globalw^{*}, \nabla F_{k} (\localw^{t}) \rangle \\
    \leq & -(F_{k}(\localw^{t}) - F_{k}(\globalw^{*})) 
    -\frac{\mu}{2} \Vert \localw^{t} - \globalw^{*} \Vert^{2}.
\end{split}
\end{equation}

By the convexity of $\Vert \cdot \Vert$ and the L-smoothness of $F_{k}(\cdot)$, we can express third term of (\ref{lamma1}) as
\begin{equation}
\label{lamma1:A3}
\begin{split}
    & \eta^{2}_{t} \Vert g_{t} \Vert ^{2} \\
    \leq & \eta^{2}_{t} \expect \Vert \nabla F_{k} (\localw^{t}) \Vert^{2} \\
    \leq & 2L \eta^{2}_{t} \expect (F_{k} (\localw^{t}) - F^{*}_{k}),
\end{split}
\end{equation}
where $F^{*}_{k}$ is the optimal model of local client $k$. Combining (\ref{lamma1})$-$ (\ref{lamma1:A3}), we have
\begin{align}
\label{lamma1:A-update}
    & \Vert \globalw^{t}- \globalw^{*} -\eta_{t}g_{t} \Vert^{2} \nonumber \\
    \leq & \Vert \globalw^{t}- \globalw^{*} \Vert^{2} + \eta_{t} \expect (\frac{1}{\eta_{t}} \Vert \globalw^{t}- \localw^{t} \Vert^{2} + \eta_{t} \Vert \nabla F_{k} (\localw^{t}) \Vert^{2}) \nonumber \\
        & - 2 \eta_{t} \expect ((F_{k}(\localw^{t}) - F_{k}(\globalw^{*})) +\frac{\mu}{2} \Vert \localw^{t} - \globalw^{*} \Vert^{2}) \nonumber \\
        &+ 2L \eta^{2}_{t} \expect (F_{k} (\localw^{t}) - F^{*}_{k}) \nonumber \\
    \leq &  (1- \mu \eta_{t}) \Vert \globalw^{t}- \globalw^{*} \Vert^{2}  + \expect \Vert \globalw^{t}- \localw^{t} \Vert^{2} \nonumber \\
            & + 4L \eta^{2}_{t} \expect (F_{k} (\localw^{t}) - F^{*}_{k}) \nonumber \\
            & - 2 \eta_{t} \expect (F_{k}(\localw^{t}) - F_{k}(\globalw^{*})),
\end{align}
where we use the L-smoothness of $F_{k}(\cdot)$ in the last inequality. And set $\gamma_{t}=2 \eta_{t} (1 - 2 L \eta_{t})$, the last two terms of (\ref{lamma1:A-update}) further become,

\begin{align}
\label{lamma1:A-update-3-4}
     & 4L \eta^{2}_{t} \expect (F_{k} (\localw^{t}) - F^{*}_{k}) \nonumber \\
    & - 2 \eta_{t} \expect (F_{k}(\localw^{t}) - F_{k}(\globalw^{*})) \nonumber \\
    = &  (4L \eta^{2}_{t} - 2 \eta_{t}) \expect (F_{k} (\localw^{t}) - F^{*}_{k}) \nonumber \\
    & - 2 \eta_{t} \expect (F_{k}(\localw^{t}) - F_{k}(\globalw^{*})) \nonumber \\
    & + 2 \eta_{t} \expect (F_{k} (\localw^{t}) - F^{*}_{k})\nonumber \\
    = & {\scriptsize -\gamma_{t} \expect (F_{k} (\localw^{t}) - F^{*}) 
     + (2 \eta_{t} - \gamma_{t})  \expect (F^{*} - F^{*}_{k})} \nonumber \\
     = & {\scriptsize -\gamma_{t} \expect (F_{k} (\localw^{t}) - F^{*}) +  4L \eta^{2}_{t} \Gamma},
\end{align}
where $\Gamma=\expect (F^{*} - F^{*}_{k})=F^{*}-\expect F^{*}_{k}$, representing the degree of data heterogeneity.

The first term of (\ref{lamma1:A-update-3-4})
\begin{align}
    & \expect ( F_{k} (\localw^{t}) - F^{*}) \nonumber \\ 
    = &  \expect (F_{k} (\localw^{t}) - F_{k}(\globalw^{t})) + \expect (F_{k}(\globalw^{t}) - F^{*}) \nonumber \\
    \geq & \expect \langle \nabla F_{k}(\globalw^{t}),  \localw^{t} - \globalw^{t})\rangle + \expect (F_{k}(\globalw^{t}) - F^{*}) \nonumber \\
    = & \expect \langle \nabla F_{k}(\globalw^{t}),  \localw^{t} - \globalw^{t})\rangle + F(\globalw^{t}) - F^{*} \nonumber \\
    \geq & -\frac{1}{2} \expect \left(\eta_{t} \Vert F_{k}(\globalw^{t}) \Vert^{2} + \frac{1}{\eta_{t}} \Vert \localw^{t} - \globalw^{t} \Vert^{2}\right) + F(\globalw^{t}) - F^{*} \nonumber \\
    \geq &  -\expect \left(\eta_{t}L (F_{k}(\globalw^{t}) - F_{k}^{*}) + \frac{1}{2\eta_{t}} \Vert \localw^{t} - \globalw^{t} \Vert^{2}\right) + F(\globalw^{t}) - F^{*},
\end{align}

where the first inequality results from the convexity of $F_{k}(\cdot)$, the second inequality from AM-GM inequality and the third inequality from L-smoothness of $F_{k}(\cdot)$.

Therefore, (\ref{lamma1:A-update-3-4}) becomes
\begin{align}
    & -\gamma_{t} \expect (F_{k} (\localw^{t}) - F^{*}) + 4L \eta^{2}_{t} \Gamma \nonumber \\
    \leq & \gamma_{t} \expect \left(\eta_{t}L (F_{k}(\globalw^{t}) - F_{k}^{*}) + \frac{1}{2\eta_{t}} \Vert \localw^{t} - \globalw^{t} \Vert^{2}\right)  \nonumber \\
    & - \gamma_{t} (F(\globalw^{t}) - F^{*}) + 4L \eta^{2}_{t} \Gamma \nonumber \\
  = & \gamma_{t} (\eta_{t}L - 1) \expect (F_{k}(\globalw^{t}) - F^{*}) \nonumber \\
  & + \frac{\gamma_{t}}{2\eta_{t}} \expect \Vert \localw^{t} - \globalw^{t} \Vert^{2} + (4L \eta^{2}_{t} + \gamma_{t} \eta_{t}L) \Gamma,
\end{align}

Since we choose $\eta_0 < \frac{1}{4}$, $\eta_{t}L - 1 < -3/4< 0$. And with $F(\globalw^{t}) - F^{*} > 0$, we have
\begin{equation*}
    \gamma_{t} (\eta_{t}L - 1) \expect (F_{k}(\globalw^{t}) - F^{*}) \leq 0,
\end{equation*}
and recall $\gamma_{t}=2 \eta_{t} (1 - 2 L \eta_{t})$, so $\frac{\gamma_{t}}{2\eta_{t}} \leq 1$ and $4L \eta^{2}_{t} + \gamma_{t} \eta_{t}L \leq 6L \eta^{2}_{t}$.
Therefore, 
\begin{equation*}
     -\gamma_{t} \expect (F_{k} (\localw^{t}) - F^{*}) + 4L \eta^{2}_{t} \Gamma \leq \expect \Vert \localw^{t} - \globalw^{t} \Vert^{2} + 6L \eta^{2}_{t} \Gamma.
\end{equation*}
Thus, (\ref{lamma1:A-update}) becomes
\begin{align}
\label{lamma1:A-final}
     & \Vert \globalw^{t}- \globalw^{*} -\eta_{t}g_{t} \Vert^{2} \nonumber \\
    \leq & (1- \mu \eta_{t}) \Vert \globalw^{t}- \globalw^{*} \Vert^{2} + 2 \expect \Vert \localw^{t} - \globalw^{t} \Vert^{2} + 6L \eta^{2}_{t} \Gamma.
\end{align}

\subsection{Bound for the divergence of weights}

To bound the weights, we assume within $E$ communication steps, there exists $t_{0}<t$, such that $t-t_{0}\leq E-1$ and $\localw^{t_{0}}=\globalw^{t_{0}}$ for all $k=1,2,\dots,m$. And we know $\eta_{t}$ is non-increasing and $\eta_{t_{0}} \leq 2\eta_{t}$. With the fact $\mathbb{E}\Vert X-\mathbb{E}X\Vert^{2} \leq \mathbb{E}\Vert X\Vert^{2}$ and Jensen inequality, we have
\begin{align}
\label{divweight}
    \mathbb{E} \expect \Vert \globalw^{t} - \localw^{t} \Vert^{2} \leq & \mathbb{E} \expect \Vert \globalw^{t_{0}} - \localw^{t} \Vert^{2} \nonumber\\
    \leq & \expect \mathbb{E} \sum_{t_{0}}^{t-1} (E-1) \eta_{t}^{2} \Vert F_{k}(\localw^{t}) \Vert^{2} \nonumber\\
    \leq & \expect \mathbb{E} \sum_{t_{0}}^{t-1} (E-1) \eta_{t_{0}}^{2} G^{2} \nonumber\\
    \leq & \expect \sum_{t_{0}}^{t-1} (E-1) \eta_{t_{0}}^{2} G^{2} \nonumber\\
    \leq & \expect (E-1)^2 \eta_{t_{0}}^{2} G^{2} \nonumber\\
    \leq & 4 \eta_{t}^{2} (E-1)^2 G^{2}.
\end{align}

\subsection{Convergence bound}
\label{analysis_of_convergence_bound}
Combining (\ref{lamma1}), (\ref{lamma1:A-final}), and (\ref{divweight}), we have
\begin{align}
    & \Vert \globalw^{t+1}-\globalw^{*}\Vert^{2} \nonumber\\
    = & \Vert \globalw^{t}- \globalw^{*} -\eta_{t}g_{t} \Vert^{2} \nonumber\\
    \leq & (1- \mu \eta_{t}) \Vert \globalw^{t}- \globalw^{*} \Vert^{2} + 2 \expect \Vert \localw^{t} - \globalw^{t} \Vert^{2} + 6L \eta^{2}_{t} \Gamma \nonumber\\
    \leq & (1- \mu \eta_{t}) \Vert \globalw^{t}- \globalw^{*} \Vert^{2} + 8\eta_{t}^{2} (E-1)^2 G^{2} + 6L \eta^{2}_{t} \Gamma.
\end{align}

Therefore, 
\begin{align}
    & \mathbb{E} \Vert \globalw^{t+1}-\globalw^{*}\Vert^{2} \\
    \leq & (1- \mu \eta_{t}) \mathbb{E}\Vert \globalw^{t}- \globalw^{*} \Vert^{2} + 8\eta_{t}^{2} (E-1)^2 G^{2} + 6L \eta^{2}_{t} \Gamma.
\end{align}

We set $\eta_{t}=\frac{\beta}{t+\gamma}$ for some $\beta > \frac{1}{\mu}$ and $\gamma > 0$, such that $\eta_{1}\leq min\{\frac{1}{\mu}, \frac{1}{4L}\}=\frac{1}{4L}$ and $\eta_{t} \leq 2 \eta_{t+E}$. We want to prove $\mathbb{E} \Vert \globalw^{t}- \globalw^{*} \Vert^{2} \leq \frac{v}{\gamma + t}$, where $v=max\{ \frac{\beta^2 B}{\beta \mu - 1}, (\gamma + 1)\mathbb{E} \Vert \globalw^{1}- \globalw^{*} \Vert^{2}\}$ and $B=8 (E-1)^{2} G^{2} + 6L\Gamma$.

Firstly, the definition of $v$ ensures that $\mathbb{E} \Vert \globalw^{1}- \globalw^{*} \Vert^{2} \leq \frac{v}{\gamma + 1}$. Assume the conclusion holds for some $t$, we have
\begin{align}
    \mathbb{E} \Vert \globalw^{t+1}-\globalw^{*}\Vert^{2} \leq & (1- \mu \eta_{t}) \mathbb{E} \Vert \globalw^{t}- \globalw^{*} \Vert^{2} + \eta_{t}^{2} B \nonumber\\
    \leq & (1 - \frac{\beta \mu}{t + \gamma})\frac{v}{t + \gamma} + \frac{\beta^{2} B}{(t + \gamma)^{2}} \nonumber\\
    = & \frac{t + \gamma - 1}{(t + \gamma)^{2}} v + [\frac{\beta^{2} B}{(t + \gamma)^{2}} - \frac{\beta \mu - 1}{(t + \gamma)^{2}} v] \nonumber\\
    \leq & \frac{v}{t + \gamma + 1}.
\end{align}

By the $L$-smoothness of $F(\cdot)$, $\mathbb{E}[F(\globalw^{t})] - F^{*} \leq \frac{L}{2} \mathbb{E} \Vert \globalw^{t}- \globalw^{*} \Vert^{2} \leq \frac{L}{2}\frac{v}{\gamma + t}$. 

Thus we have
\begin{align*}
    & \mathop{\mathbb{E}}[F(\weight_{T})] - F^{*} \\
    \leq & {\frac{2}{\mu^{2}}} \cdot \frac{L}{\gamma +T}(6L\Gamma + 8(E-1)^{2}G^2 + \frac{\mu^{2}}{4} \mathbb{E}\Vert \globalw^{1}-\globalw^{*}\Vert^{2}).
\end{align*}

\section{Convergence Analysis for Partial Device Participation}
\label{convergence_partial}

With partial device participation, each time there are $m=C \cdot M$ clients involved in aggregation. Suppose the aggregated global model with full model participation is $\fullw=\sum_{k=1}^{M} p_{k} \localw$, which is different from the aggregated global model with partial model participation $\globalw = \expect \localw$. 

\begin{assumption}
    \label{assp:equal_prob}
    Suppose the weight $p_k$ of each device is the same, which is, $p_1=p_2=\cdots=p_M=\frac{1}{M}$.
\end{assumption}

As each sampling distribution is identical, and Assumption~\ref{assp:equal_prob} holds, we have unbiased sampling scheme,
\begin{align*}
    \mathop{\mathbb{E}}(\globalw) = \fullw.
\end{align*}

And its proof is as follows,
\begin{align*}
    & \mathop{\mathbb{E}}(\globalw) = \mathop{\mathbb{E}} (\expect \localw) = \expect \mathop{\mathbb{E}} (\localw)\\
     = & \expect \sum_{k=1}^{M} q_{k} \localw = \sum_{k=1}^{M} q_{k} \localw = \fullw.
\end{align*}

\subsection{Analysis on consecutive steps}
Similar to the previous analysis, to bound the expectation of the global objective function at time $T$ from its optimal value, we first consider analyzing the global weight from the optimal weights by calculating single-step SGD:
\begin{align}
\label{lamma1:one_step}
    \Vert \globalw^{t+1}- \globalw^{*}\Vert^{2} & = \Vert \globalw^{t+1}- \fullw^{t+1} + \fullw^{t+1} - \globalw^{*}\Vert^{2} \nonumber \\
    & = \Vert \globalw^{t+1}- \fullw^{t+1} \Vert^{2} + \Vert \fullw^{t+1}- \globalw^{*} \Vert^{2} \nonumber \\
    & + 2\langle \fullw^{t+1}- \globalw^{*}, \globalw^{t+1}- \fullw^{t+1} \rangle.
\end{align}

Once taking the expectation over selected devices, we have,
\begin{align}
\label{lamma1:one_step_exp}
    \mathbb{E} \Vert \globalw^{t+1}- \globalw^{*}\Vert^{2} & = \mathbb{E}\Vert \globalw^{t+1}- \fullw^{t+1} \Vert^{2} + \mathbb{E}\Vert \fullw^{t+1}- \globalw^{*} \Vert^{2}.
\end{align}

Due to the unbiased sampling, the expectation of the third term of~(\ref{lamma1:one_step}) is 0. Based on previous analysis, the second term of~(\ref{lamma1:one_step_exp}) becomes,
\begin{align}
    \mathbb{E} \Vert \fullw^{t+1}-\globalw^{*}\Vert^{2} \leq (1- \mu \eta_{t}) \mathbb{E} \Vert \fullw^{t}- \globalw^{*} \Vert^{2} + \eta_{t}^{2} B.
\end{align}

And~(\ref{lamma1:one_step_exp}) becomes,
\begin{align}
    \label{lamma:partial_selected}
    & \mathbb{E} \Vert \globalw^{t+1}- \globalw^{*}\Vert^{2} \nonumber \\
    = & \mathbb{E}\Vert \globalw^{t+1}- \fullw^{t+1} \Vert^{2} + \mathbb{E}\Vert \fullw^{t+1}- \globalw^{*} \Vert^{2} \nonumber \\
    \leq & \mathbb{E}\Vert \globalw^{t+1}- \fullw^{t+1} \Vert^{2} + (1- \mu \eta_{t}) \mathbb{E} \Vert \fullw^{t}- \globalw^{*} \Vert^{2} + \eta_{t}^{2} B.
\end{align}

\subsection{Bound for the variance of $\globalw^t$}
Based on the Assumption~\ref{assp:equal_prob}, we have $\fullw^{t+1} = \frac{1}{M} \sum_{k=1}^M \localw^{t+1}$ and $\globalw^{t+1} = \frac{1}{m} \sum_{k=1}^m \localw^{t+1}$, where $m=C\cdot M$ is the number of selected clients. And the set of selected clients is denoted as $\mathcal{S}_{t}$. In this case, the first term of~(\ref{lamma1:one_step_exp}) becomes,

\begin{align}
    \label{bound_for_variance}
    & \mathbb{E}_{\mathcal{S}_{t+1}}\Vert \globalw^{t+1}- \fullw^{t+1} \Vert^{2} \nonumber \\
    = & \mathbb{E}_{\mathcal{S}_{t+1}} \Vert \frac{1}{m} \sum_{k=1}^m \localw^{t+1}- \fullw^{t+1} \Vert^{2} \nonumber \\
    = & \frac{1}{m^2} \Vert \sum_{k=1}^M \mathbb{I}{(k \in \mathcal{S}_{t+1})} (\localw^{t+1}- \fullw^{t+1}) \Vert^{2} \nonumber \\
    = & \frac{1}{mM} \sum_{k=1}^M \Vert (\localw^{t+1}- \fullw^{t+1}) \Vert^{2} \nonumber\\
    & +  \frac{m-1}{mM(M-1)} \sum_{k_i \neq k_j}^M \langle \mathbf{w}_{k_i}^{t+1} -\fullw^{t+1}, \mathbf{w}_{k_j}^{t+1} -\fullw^{t+1}  \rangle \nonumber \\
    = & (\frac{1}{mM} - \frac{m-1}{mM(M-1)}) \sum_{k=1}^M \Vert (\localw^{t+1}- \fullw^{t+1}) \Vert^{2},
\end{align}
where in the last equality we use $\sum_{k=1}^M \Vert (\localw^{t+1}- \fullw^{t+1}) \Vert^{2} + \sum_{k_i \neq k_j}^M \langle \mathbf{w}_{k_i}^{t+1} -\fullw^{t+1}, \mathbf{w}_{k_j}^{t+1} -\fullw^{t+1}  \rangle = \Vert \sum_{k=1}^M(\localw^{t+1}- \fullw^{t+1}) \Vert^{2} = 0$.

Therefore, 
\begin{align}
    \label{bound_for_variance_2}
    & \mathbb{E}\Vert \globalw^{t+1}- \fullw^{t+1} \Vert^{2} \nonumber \\
 = & (\frac{1}{mM} - \frac{m-1}{mM(M-1)}) \mathbb{E}\sum_{k=1}^M \Vert (\localw^{t+1}- \fullw^{t+1}) \Vert^{2} \nonumber \\
    \leq & (\frac{1}{m} - \frac{m-1}{m(M-1)}) \mathbb{E}\sum_{k=1}^M \Vert \frac{1}{M} (\localw^{t+1}- \globalw^{t_0}) \Vert^{2} \nonumber \\
    \leq & (\frac{1}{m} - \frac{m-1}{m(M-1)}) 4 \eta_{t}^{2} (E-1)^2 G^{2} \nonumber \\
    = & 4 \frac{M-m}{m(M-1)} \eta_{t}^{2} (E-1)^2 G^{2},
\end{align}
where the last inequality is based on~(\ref{divweight}).

\paragraph{Influences of latent clients.}
As we perform the proposed procedure, the aggregated model is merged with additional latent clients, which is, $\globalw'^{,t+1} = \frac{1}{m^2} \sum_{k=1}^m \localw^{t+1} + \frac{1}{m^2} \sum_{k_i\neq k_j}^m \mathbf{w}_{k_i,k_j}^{t+1}$. Based on Section~\ref{theory:variance}, we have $\mathbb{E} \Vert \globalw'^{,t+1} - \fullw^{t+1}\Vert^{2} \leq \mathbb{E} \Vert \globalw^{t+1} - \fullw^{t+1}\Vert^{2}$, which means lower variance of $\globalw^t$ with our method. Compare it with~(\ref{bound_for_variance}), we have,

\begin{align}
    & \mathbb{E}_{\mathcal{S}'_{t+1}}\Vert \globalw'^{,t+1}- \fullw^{t+1} \Vert^{2} \\
    = & \mathbb{E}_{\mathcal{S}'_{t+1}} \Vert \frac{1}{m^2} (\sum_{k=1}^m \localw^{t+1} + \sum_{k_i\neq k_j}^m \mathbf{w}_{k_i,k_j}^{t+1})- \fullw^{t+1} \Vert^{2} \nonumber \\
    = & (\frac{1}{m^2 M^2} - \frac{m^2-1}{m^2 M^2(M^2-1)}) \sum_{k_i=1}^M \sum_{k_j=1}^M \Vert (\mathbf{w}_{k_i,k_j}^{t+1}- \fullw^{t+1}) \Vert^{2},
\end{align}
where $\mathcal{S}'_{t}$ denotes the set of selected clients and reconstructed latent clients. And it is straightforward to calculate that as $m>1, M>1$, $\frac{1}{m^2 M^2} - \frac{m^2-1}{m^2 M^2(M^2-1)} < \frac{1}{mM} - \frac{m-1}{mM(M-1)}$. It is aligned with the low variance argument of $\globalw'^{,t+1}$.

\subsection{Convergence bound}
Combine~(\ref{lamma:partial_selected}) and~(\ref{bound_for_variance_2}), we have,
\begin{align}
    \label{lamma:partial_selected_2}
    \mathbb{E} \Vert \globalw^{t+1}- \globalw^{*}\Vert^{2} & = \mathbb{E}\Vert \globalw^{t+1}- \fullw^{t+1} \Vert^{2} + \mathbb{E}\Vert \fullw^{t+1}- \globalw^{*} \Vert^{2} \nonumber \\
    & \leq (1- \mu \eta_{t}) \mathbb{E} \Vert \fullw^{t}- \globalw^{*} \Vert^{2} + \eta_{t}^{2} (B+D),
\end{align}
where $D = 4 \frac{M-m}{m(M-1)} (E-1)^2 G^{2}$ is the upper bound of $\frac{1}{\eta_{t}^{2}} \mathbb{E}_{\mathcal{S}_{t+1}}\Vert \globalw^{t+1}- \fullw^{t+1} \Vert^{2}$. While with our method, $D = 4 \frac{M^2-m^2}{m^2 M^2(M^2-1)} (E-1)^2 G^{2}$ is the upper bound of $\frac{1}{\eta_{t}^{2}} \mathbb{E}_{\mathcal{S}'_{t+1}}\Vert \globalw'^{t+1}- \fullw^{t+1} \Vert^{2}$.

With the same form as in Section~\ref{analysis_of_convergence_bound}, we can prove $\mathbb{E} \Vert \globalw^{t}- \globalw^{*} \Vert^{2} \leq \frac{v}{\gamma + t}$, where $v=max\{ \frac{\beta^2 (B+D)}{\beta \mu - 1}, (\gamma + 1)\mathbb{E} \Vert \globalw^{1}- \globalw^{*} \Vert^{2}\}$. Specifically, if we choose $\beta=2/\mu$ we have
\begin{align*}
    \mathop{\mathbb{E}}[F(\weight_{T})] - F^{*}
    \leq {\frac{2}{\mu^{2}}} \cdot \frac{L}{\gamma +T}(B+D + \frac{\mu^{2}}{4} \mathbb{E}\Vert \globalw^{1}-\globalw^{*}\Vert^{2}).
\end{align*}

\section{Fast/Slow Communication Protocol}
\label{sec:fast_slow}
Compared with atom coefficients $\coeff \in \mathbb{R}^{m_a\times c'\times c}$, the filter atoms $\atom \in \mathbb{R}^{k_a\times k_a \times m_a}$ have significantly fewer parameters since $k_a \times k_a \ll c'\times c$, a few hundred of parameters typically. To incorporate this finding, we further adopt a fast/slow communication protocol, which prioritizes the transmission of local knowledge, \textit{i.e.}, filter atoms, over atom coefficients to minimize communication costs. More precisely, we introduce a parameter $\beta$, that determines the frequency of atom coefficient communication. For instance, if $\beta=1/10$, the atom coefficients are communicated and updated once every ten rounds.

With this, the parameters of the model can be aggregated as follows,
\begin{equation}
    \begin{bmatrix} 
        \coeff^{t+1} \\ 
        \atom^{t+1} \\ 
        \param_{h}^{t+1} 
    \end{bmatrix} 
    \xleftarrow{} 
    \begin{bmatrix} 
        \sum_{k=1}^{m} \frac{n_k}{n} \coeff^{t+1}_k \mathbf{\delta}_{\{\beta t \in \mathbb{N}\}} + \coeff^{t} \mathbf{\delta}_{\{\beta t \notin \mathbb{N}\}}\\ 
        \sum_{k=1}^{m} \frac{n_k}{n} \atom^{t+1}_k  \\ 
        \sum_{k=1}^{m} \frac{n_k}{n}  \param_{k, h}^{t+1} 
    \end{bmatrix},
    \label{eq:beta_dcf_weight_coeff_update}
\end{equation}
where $\mathbf{\delta}_{\{\beta t \in \mathbb{N}\}}$ is the indicator which equals to 1 only when $\beta t$ is an integer. The algorithm is summarized in Appendix Algorithm~\ref{alg:algorithm_fast_slow}.

\paragraph{Reduction in Communication Overhead.} Suppose the number of parameters in classification heads is $l_1 \times l_2$, \textit{i.e.}, $\param_{h} \in \mathbb{R}^{l_1 \times l_2}$.
the communication complexity of transmitting atoms and classification head is $\mathcal{O}(m_a \cdot k_a^2 + l_1 \cdot l_2)$, while transmitting the entire model needs atoms, coefficients and classification heads, which is of $\mathcal{O}(m_a \cdot k_a^2 + c \cdot c' \cdot m_a + l_1 \cdot l_2)$. 
By utilizing the fast/slow communication protocol, the reduction rate of the total communicated parameters is expressed as $\frac{\beta(m_a \cdot k_a^2 + c \cdot c' \cdot m_a + l_1 \cdot l_2) + (1-\beta) (m_a \cdot k_a^2 + l_1 \cdot l_2)}{(m_a \cdot k_a^2 + c \cdot c' \cdot m_a + l_1 \cdot l_2) } \approx \beta$, compared to communicating the entire model. 

And the training procedure of fast/slow communication protocol is summarized in Algorithm~\ref{alg:algorithm_fast_slow}.  $M$ is the total number of clients, and $C$ is the ratio of selected clients at each communication round. $T$ is the total communication round. $\globalw^{0}$ are the initialized parameters, and $\eta_0$ is the initial learning rate. $B, E$ are the local training batch size and local training epoch. Different from Algorithm~\ref{alg:algorithm}, the fast/slow communication protocol only transmits atom coefficients $\coeff$ every once of $1/\beta$ round to decrease the communication overhead.

\begin{algorithm}[]
\caption{Fast/Slow Communication Protocol}
\label{alg:algorithm_fast_slow}
\textbf{Input}: $M, T, \beta, \globalw^{0}, \eta_0, B, C, E$ \\
\textbf{Server:}
\begin{algorithmic} 
\FOR{$t = 0, 1, 2, \dots, T-1$}
\item $m \xleftarrow{} max(C\cdot M,1)$
\item $S_{t} \xleftarrow{} $Random set of $m$ clients
\FOR{k $\in S_{t}$}
\IF{$t \beta \in \mathbb{N}$}
    \STATE $\coeff^{t+1}_{k}, \atom^{t+1}_{k}, \param_{k, h}^{t+1} \xleftarrow{} Client(k, \coeff^{t}, \atom^{t}, \param_{h}^{t})$
\ELSE
    \STATE $\atom^{t+1}_{k}, \param_{k, h}^{t+1} \xleftarrow{} Client(k, \coeff^{t}, \atom^{t}, \param_{h}^{t})$
\ENDIF
\ENDFOR
\STATE{\emph{// weight aggregation}}
\IF{$t \beta \in \mathbb{N}$}
    \STATE $\coeff^{t+1} \xleftarrow{} \sum_{k=1}^{m} \frac{n_k}{n} \coeff^{t+1}_k $
\ELSE
    \STATE $\coeff^{t+1} \xleftarrow{} \coeff^{t} $
\ENDIF
\item $\atom^{t+1} \xleftarrow{} \sum_{k=1}^{m} \frac{n_k}{n} \atom^{t+1}_k $
\item $\param_{h}^{t+1} \xleftarrow{} \sum_{k=1}^{m} \frac{n_k}{n}  \param_{k, h}^{t+1} $

\STATE{\emph{// re-construct global model}}
\item $\param_{\extractor}^{t+1} \xleftarrow{} \coeff^{t+1} \times \atom^{t+1}$
\ENDFOR\\

\end{algorithmic}
\textbf{Client:}
\begin{algorithmic}
\FOR{$i=0,1,2,\dots, E-1$}
\FOR{$b \in B$}
\item $\coeff^{t+1}_{k} \xleftarrow{} \coeff^{t}_k - \eta_t \nabla_{\coeff^{t}_k} F_k$
\item $\atom^{t+1}_{k} \xleftarrow{} \atom^{t}_k - \eta_t \nabla_{\atom^{t}_k} F_k$
\item $\param_{k, h}^{t+1} \xleftarrow{} \param_{k, h}^{t} - \eta_t \nabla_{\param_{k, h}^{t}} F_k$
\ENDFOR
\ENDFOR
\IF{$t \beta \in \mathbb{N}$}
    \STATE \textbf{return} $\coeff^{t+1}_{k}, \atom^{t+1}_{k}, \param_{k, h}^{t+1}$
\ELSE
    \STATE \textbf{return} $\atom^{t+1}_{k}, \param_{k, h}^{t+1}$
\ENDIF
\end{algorithmic}
\end{algorithm}

\subsection{Convergence Analysis of Fast/Slow Communication Protocol}
\paragraph{The weigh divergence bound for fast/slow communication protocol.} Since the fast/slow communication protocol transmits the atom coefficients $\coeff_k$ once every $1/\beta$ round, within $E/\beta$ communication steps, there exists $t_{0}<t$, such that $t-t_{0}\leq E/\beta-1$ and $\localw^{t_{0}}=\globalw^{t_{0}}$ for all $k=1,2,\dots,m$. Similar to (\ref{divweight}), we have the bound for the divergence of weights,
\begin{align}
\label{divweight_fast_slow}
    \mathbb{E} \expect \Vert \globalw^{t} - \localw^{t} \Vert^{2} 
    \leq & 4 \eta_{t}^{2} (E/\beta-1)^2 G^{2}.
\end{align}


The convergence outcome of the fast/slow communication protocol, which involves transmitting only filter atoms every round while the entire model including atom coefficients every $1/\beta$ round, can be expressed as follows:

\begin{theorem}
    Let Assumptions~\ref{assp:lsmooth} to~\ref{assp:gradbound}  hold and $L,\mu, G$ be defined therein, with filter atoms transmitted every round and the entire model communicated every $1/\beta$ round. Choose $\gamma=max\{8\frac{L}{\mu},E\}$, and $\eta_{t}=\frac{2}{\mu(\gamma + t)}$. Let $F^{*}$ and $F_{k}^{*}$ be the minimum value of global model $F$ and each local model $F_{k}$ respectively, then:
    \begin{align*}
        & \mathop{\mathbb{E}}[F(\globalw^{T})] - F^{*}  \\
        \leq & {\frac{2}{\mu^{2}}} \cdot \frac{L}{\gamma +T}(6L\Gamma + 8(E/\beta-1)^2G^2 + \frac{\mu^{2}}{4} \mathbb{E}\Vert \globalw^{0}-\globalw^{*}\Vert^{2}).
    \end{align*}
\end{theorem}

The convergence analysis of the fast/slow protocol results in a slightly looser bound compared to Theorem~\ref{thm:converge}, as $8(E/\beta-1)^2G^2 > 8(E-1)^2G^2$ with $\beta < 1$. 
It means the fast/slow protocol takes more time to reach the same amount of convergence bound than the regular communication protocol. 
However, the convergence rate is still approximately $O(\frac{1}{T})$. And because of its lower communication cost at each round, the overall communication cost of fast/slow protocol is less than the regular communication protocol, further empirically validated in Section~\ref{sec:comm_eff}.
Appendix~\ref{theory:convergence} shows a formal proof of this theorem.

\section{Experimental Settings}
\label{exp:setting}
\paragraph{Dataset partitions.}
Suppose a dataset contains $N$ training data with $K$ classes, and the data are randomly distributed to $M$ clients with $K'$ classes in each client, which is the case $(M, K')$. To partition the data, we adopt the data partition rule outlined in \citep{mcmahan2017communication}, where the dataset is divided into $S_n=M*K'$ shards. Each shard contains ${N}/{S_n}$ images from a single class. Each client stores $K'$ shards locally.

\paragraph{Training time.}
Each model is trained on Nvidia RTX A5000 for 200 communication rounds. With 100 clients, the training time for Ditto~\citep{li2021ditto} and FedPac~\citep{fedpac} is $3.2$ hours, while other methods take about $1$ hour. With 1000 clients, the training time for Ditto and FedPac is over $35$ hours, while other methods take about $19.7$ hours.

\paragraph{Hyper-Parameters.}
Our implementation adapts codebase from~\citep{collins2021exploiting, fedpac}. The optimizer of all the methods is SGD with a learning rate of 0.01 and momentum of 0.9. The local batch size is 10 and the local training epoch is 10. We set FedDyn’s hyper-parameter as 0.01. For Ditto we used $\lambda=0.75$ for all cases. For FedProx we use $\mu=0.0001$. For FedRep, we follow the same setting in~\citep{collins2021exploiting}. For each local update, FedRep executes 10 local epochs to train the local head, followed by 1 epoch for the representation.

\section{Extra Experiments}
\label{sec:extra_exp}

\paragraph{Influence of different $m_a$}
The corresponding results are presented in Table~\ref{tab:different_ma}. 
Larger values of $m_a$ correspond to higher accuracy. Smaller values of $m_a$ lead to fewer involved parameters, thus less computational resource required for training and less communication overhead.

For simpler datasets, such as CIFAR, the differences in accuracy among $m_a=6,9,12$ are relatively small. However, in more challenging tasks, such as Tiny-ImageNet, employing a greater number of filter atoms $m_a=9,12$ results in a more substantial accuracy gap compared to $m_a=3,6$. Therefore, it more preferable to use smaller $m_a$ in experiments with CIFAR datasets but a larger $m_a$ in experiments with Tiny-ImageNet. 

\begin{table}
\caption{The test accuracy for different numbers of filter atoms $m_a$.}
\centering
  \scalebox{0.65}{
  \label{tab:different_ma}
  \centering
   \begin{tabular}{l|cccccc}
\toprule
                           & \multicolumn{2}{c}{CIFAR10}    & \multicolumn{2}{c}{CIFAR100}    & \multicolumn{2}{c}{Tiny-ImageNet} \\ \cline{2-7} 
                           & (100,2)       & (100,5)        & (100,5)        & (100,20)       & (1000,20)       & (1000,50)       \\ \midrule
$m_a=12$     & 94.25         & 87.08 & 84.66          & 63.67          & 42.25           & 23.81           \\
$m_a=9$     & 94.14         & 87.49 & 84.28          & 63.28          & 41.57           & 24.32           \\
$m_a=6$     & 94.29         & 85.92 & 83.38          & 62.71          & 39.4           & 21.5           \\
$m_a=3$     & 91.98         & 81.66 & 80.27          & 57.68          & 38.51           & 20.91           \\ \bottomrule
\end{tabular}
 }
\end{table}

\end{document}